\documentclass[suppldata]{interact}

\usepackage[colorlinks,bookmarksopen,bookmarksnumbered,citecolor=blue,urlcolor=blue,linkcolor = blue]{hyperref}

\usepackage{epstopdf}% To incorporate .eps illustrations using PDFLaTeX, etc.
\usepackage{subfigure}% Support for small, `sub' figures and tables

\usepackage{natbib}% Citation support using natbib.sty
\bibpunct[, ]{(}{)}{,}{a}{}{,}% Citation support using natbib.sty
\renewcommand\bibfont{\fontsize{10}{12}\selectfont}% Bibliography support using natbib.sty

\theoremstyle{plain}% Theorem-like structures
\newtheorem{theorem}{Theorem}[section]

\theoremstyle{definition}
\newtheorem{definition}[theorem]{Definition}

\theoremstyle{remark}

\usepackage{amsmath} 
\usepackage{booktabs}
\usepackage{multirow}
\usepackage{float}
\usepackage[compact]{titlesec}
\usepackage{color,soul}

\usepackage[pagewise]{lineno}

\usepackage{cleveref}

\usepackage{geometry}
\begin{document}

\articletype{RESEARCH ARTICLE}

\title{Quantifying geographic domain shift to decouple the geospatial transferability of human mobility flow generation models}

\author{
\name{Zhiyong Zhou \textsuperscript{a,b}, Song Gao \textsuperscript{a,*}\thanks{*A preprint draft and the final version will be available on the Annals of AAG; *Corresponding Author, Song Gao, Email: song.gao@wisc.edu}, Qianheng Zhang \textsuperscript{a}, Feng Zhang \textsuperscript{b}, Zhenhong Du \textsuperscript{b}}  
\affil{\textsuperscript{a}GeoDS Lab, Department of Geography, University of Wisconsin-Madison, Madison, WI, USA\\
\textsuperscript{b}School of Earth Sciences \& Zhejiang Key Laboratory of Geographic Information Science, Zhejiang University, Hangzhou, China}}

\maketitle

\begin{abstract}
Human mobility serves as an essential proxy for understanding social, economic, and environmental dynamics in urban systems. However, since human mobility data are often scarce due to high collection costs and privacy concerns, generating them from limited mobility observations or auxiliary data sources is needed. Geospatial transferability, which measures a model's capability in a new location or unseen region, is a critical dimension for comparing different human mobility generation models. However, few studies have studied the intrinsic characteristics of geospatial transferability. To this end, this study systematically investigates the geospatial transferability of four representative human mobility generation models using a large-scale benchmark dataset of census tract–level commuting flows across 2,265 counties in the United States. Inspired by the domain adaptation theory in machine learning, we introduce geographic domain shift to describe the intrinsic differences in geographic feature distributions and spatial structures between source and target regions, which may jointly affect model transferability. Moreover, we propose two metrics, mutual information and spatial shift, to quantify the geographic domain shift. To examine their associations with model transferability, we employ linear mixed-effects regression to analyze the associations between geographic domain shifts and transferability. Our results reveal substantial spatial heterogeneity and asymmetry in transfer performance across regions. Both information shift and spatial shift exhibit statistically significant and complementary explanatory power. This indicates that geospatial transferability depends not only on model design but also on intrinsic geographic differences. These findings provide a novel methodological framework for evaluating and improving the geospatial transferability of human mobility generation models and support more robust and fair human mobility data synthesis across diverse regions. It also offers insights on spatial transferability for GeoAI model development.
\end{abstract}


\section{Introduction}
Human mobility is an essential proxy for spatial-social dynamics embedded in urban and transportation systems \citep{pappalardo2023future, xu2016another,xu2023urban,zhu2026gravity}, including social segregation \citep{nilforoshan2023human, xu2025using}, socioeconomic status \citep{xu2018human,wang2024infrequent, yabe2025behaviour}, tourism~\citep{xu2021tourism}, pandemic responses \citep{hou2021intracounty,huang2022staying,noi2022assessing,santana2023covid}, and environmental sustainability \citep{zheng2024impacts}. However, human mobility data are often scarce and limited in geographic coverage due to the high costs associated with the deployment of localization infrastructure and the significant privacy concerns about the collection and access of location data, which constrains comprehensive analyses of social dynamics. Therefore, synthesizing or generating realistic human mobility data has become as a crucial approach for mitigating data scarcity or biases and compensating for missing or inaccessible mobility records \citep{luca2021survey}.

Although physical and statistical models have long been used to characterize individual- and population-level mechanisms of human mobility \citep{gonzalez2008understanding, schlapfer2021universal, boucherie2025decoupling, chen2025addressing}, data-driven machine learning approaches have demonstrated stronger predictive and generative capabilities, even when human movement observations are sparse or entirely unavailable. These models leverage auxiliary static geographic information, such as satellite imagery, demographic attributes, and points of interest (POI), to infer human mobility patterns \citep{xu2025predicting, rong2023goddag, rong2025large, simini2021deep}. However, because training data are typically drawn from a limited set of geographic areas, the geospatial transferability of human mobility generation models has become a critical concern due to the underlying data bias and spatial heterogeneity of physical and social environments~\citep{luca2021survey,goodchild2021replication}. 

To address this issue, existing studies have incorporated prior knowledge of human mobility behavior, such as gravity law, visitation law, and long-tail distribution of mobility flows, into deep learning models. These approaches have been shown to improve cross-region transfer learning performance when models are evaluated on datasets from regions different from the training area \citep{simini2021deep,schlapfer2021universal,rong2023origin, wang2025deep, zhao2025predicting,yang2026transferable}. To date, however, assessments of model transferability typically assume that the target regions are inherently different from the training region. While this assumption facilitates comparisons among different human mobility generation models under a fixed evaluation setting, it overlooks the degree of intrinsic geographic differences between training and testing regions. In practice, a transferred geographic area is not necessarily different from the training area, as geographic difference or dissimilarity can be shaped by spatial proximity \citep{tobler1970computer}, socioeconomic status \citep{goodchild2004validity}, physical environments \citep{zhang2024urban, yan2024quantifying}, and environmental configurations~\citep{zhu2018spatial}. In the context of human mobility flows, several studies have empirically identified regional effects and spatial heterogeneity of human mobility flows \citep{xu2023spatial, chen2025addressing, long2025data}. 
 
By overlooking such intrinsic geographic differences, existing evaluations leave unresolved how difficult it is to transfer a human mobility generation model from one region to another. This limitation can further bias the generation of human mobility data across different geographic areas, a challenge increasingly observed in geospatial artificial intelligence (GeoAI) models more broadly \citep{gao2023handbook,lou2025geoxcp, hou2025transferred, wang2025geobs, zhang2026city}. Hence, we argue that the geospatial transferability of human mobility generation models is determined not only by model architectures, but also by the intrinsic geographic differences between the source (training) region and the target (testing or transferred) region. 

In this study, we focus on the generation of human travel origin-destination (OD) flow data, which represent aggregated individual movements between census geographic origin–destination pairs \citep{luca2021survey}, with four representative mobility generation models, including  \textit{DeepGravity} \citep{simini2021deep}, boosting regression tree (GBRT) \citep{robinson2018machine}, random forest (RF) \citep{pourebrahim2019trip}, and geo-contextual multitask embedding learning (GMEL) \citep{liu2020learning}. The models are trained with geographic features and OD flows in a state. Subsequently, they are transferred to each county of the remaining states to evaluate their \textit{geospatial transferability}. Meanwhile,  we propose two domain shift metrics, \textit{mutual information (MI) shift} and \textit{Moran's I spatial shift} (i.e., Moran shift for simplicity in the following text), to quantify the intrinsic differences in geographic feature distribution and spatial structure between a source area and a target area. Inspired by domain adaptation and generalization theories in transfer learning \citep{zhuang2020comprehensive, liu2023towards}, we refer to these differences as \textit{geographic domain shift}. Furthermore, we analyze the associations between geographic domain shift and model transferability to validate the proposed metrics and to understand how intrinsic differences in geographic features influence the transfer performance of human mobility OD flow generation models.

Notably, rather than developing a new geographically transferable deep learning model for generating human mobility OD flows, this study focuses on characterizing the intrinsic properties of geographic features in the training (source) and testing (target) regions \textit{prior to model training}. By using a large-scale benchmark dataset of census tract–level commuting OD flows within each of 2,265 counties in the United States \citep{rong2025large}, we demonstrate the feasibility and effectiveness of the proposed geographic shift metrics. These metrics will enable the adaptive selection of suitable training areas for transferring a human mobility generation model to other geographic areas. Moreover, they are promising for guiding the optimization of human mobility generation models toward improved geospatial transferability and fairness in synthetic human mobility data. The primary contributions of this study are three-fold:
\begin{itemize}

    \item We conceptualize geographic domain shift and propose quantitative measures to characterize the intrinsic differences between source and target geographic domains. These measures provide a basis for interpreting geospatial transferability and may inform future model optimization and transfer learning strategies.

    \item We demonstrate that the geospatial transferability of multiple human mobility generation models varies substantially across target geographic areas. This finding highlights the need for more transparent reporting of transfer learning settings and for quantitative descriptions of how difficult a given geographic area can be transferred to or from.
    
    \item We employ linear mixed-effects regression to analyze the associations between the geographic domain shift metrics and model transferability. Our analysis results reveal substantial spatial heterogeneity and asymmetry in cross-region transferability.
\end{itemize}

\section{Related work}

\subsection{Geospatial transferability of human mobility generation models}
Human mobility generation can basically be grouped into individual mobility trajectory generation and aggregated mobility flow generation \citep{luca2021survey}. In this study, we focus on the latter, which assigns inflow and outflow volumes to each pair of origin-destination (OD) geographic areas, referred to as OD flow generation. The geospatial transferability of a mobility generation model describes its ability to generate accurate mobility flows in a new or previously unseen geographic area. It is commonly based on OD flow distribution similarity measures, such as common part of commuters (CPC) and Jensen-Shannon Divergence (JSD), and magnitude-based errors like root mean squared errors (RMSE) of OD flows \citep{barbosa2018human, rong2025large}.

Existing studies on OD flow model transferability are largely \textit{model-centric}, focusing on developing new mobility generation architectures followed by post-hoc evaluations of their transfer performance. For example, \citet{simini2021deep} evaluated the transferability of the DeepGravity model across cities using a \textit{leave-one-city-out} strategy. A growing body of work further explores spatial transfer learning for human mobility generation by learning domain-invariant representations that generalize across regions. These approaches include hierarchical mobility knowledge transfer \citep{he2020human}, embedding the learning of novel data \citep{jiang2021transfer}, adversarial training \citep{rong2023goddag, yuan2025learning}, meta-learning \citep{wang2024cola}, and large language models \citep{yu2024harnessing}. Although these studies validate their models using unseen geographic areas, they rarely investigate why the same model exhibits varying levels of geospatial transferability across different target regions. In particular, little attention has been paid to how intrinsic geographic differences between regions influence the transfer performance of OD flow generation models.

\subsection{Quantification of domain shift}
As implied by prior work on transfer learning, a primary challenge hindering the geospatial transferability of deep learning models, which is independent of tasks, is \textit{domain shift} across geographic areas \citep{pan2009survey, moreno2012unifying, liu2023towards}. The domain shift typically encompasses both the covariate shift and the concept shift. The former refers to changes in the distribution of input features \citep{cai2025diagnosing}, while the latter denotes changes in the conditional distribution of labels given the input features \citep{garg2020unified} or changes in concept's meanings over time \citep{shi2025defining}. Since this study focuses on transferring a human mobility generation model to previously unseen geographic areas, where the label distribution is unknown, we mainly investigate the quantification of \textit{covariate shift}. 

In the field of transfer learning, many distance measures have been proposed to quantify the covariate shift \citep{farahani2021brief}, such as mutual information \citep{blitzer2007biographies}, maximum mean discrepancy (MMD; \citealp*{gretton2006kernel}), Wasserstein distance \citep{panaretos2019statistical}, correlation alignment (CORAL; \citealp*{sun2017correlation}), and Kullback-Leibler (KL) divergence \citep{kullback1951information}. They primarily characterize discrepancies in the statistical distributions of covariates or feature representations.

From a geographic perspective, however, domain shift is often linked to \textit{geographic similarity} \citep{goodchild2004validity, mcintosh2005assessing, zhu2018spatial}. Methodologically, many geographic similarity metrics are derived from the statistical properties of covariate distributions \citep{zhu2015predictive, zhao2025multivariate}. Beyond purely statistical similarity, researchers have also modeled the \textit{semantic similarity} of geographic features to distinguish between geographic domains \citep{schwering2008approaches, janowicz2011semantics}. Nevertheless, these geographic similarity metrics generally do not account for the spatial distribution of geographic features and the associated structural differences in geographic phenomena across areas. Such spatial structure shifts are fundamental to geographic processes and can be explicitly characterized using classical spatial statistics, such as Moran’s I \citep{moran1950notes, lee2017extending}.
% for comparable clustering or dispersion levels of geographic phenomena

\subsection{Research gaps}
Existing research on the geospatial transferability of human mobility generation models has largely focused on post-hoc evaluations to validate newly proposed models. In contrast, little attention has been paid to the intrinsic characteristics of cross-regional differences, such as geographic features and spatial structures, that underlie observed variations in transfer performance. Moreover, few studies have systematically examined how model transferability relates to the intrinsic characteristics of input data.

Meanwhile, although research on domain adaptation in machine learning has produced several methods for quantifying domain shifts, including geographic domains, they rarely account for changes in the spatial structures underlying geographic features or covariates. Therefore, a quantification framework for geographic domain shift, that is, capturing both distributional shifts and spatial-structure shifts, remains underexplored. This gap further constrains our understanding of the potential influence of intrinsic input data differences across locations and regions on the geospatial transferability of human mobility generation models.

\section{Data and methods}

\subsection{OD mobility flow dataset}
\begin{figure}[!ht]
    \centering
    \includegraphics[width=0.95\textwidth]{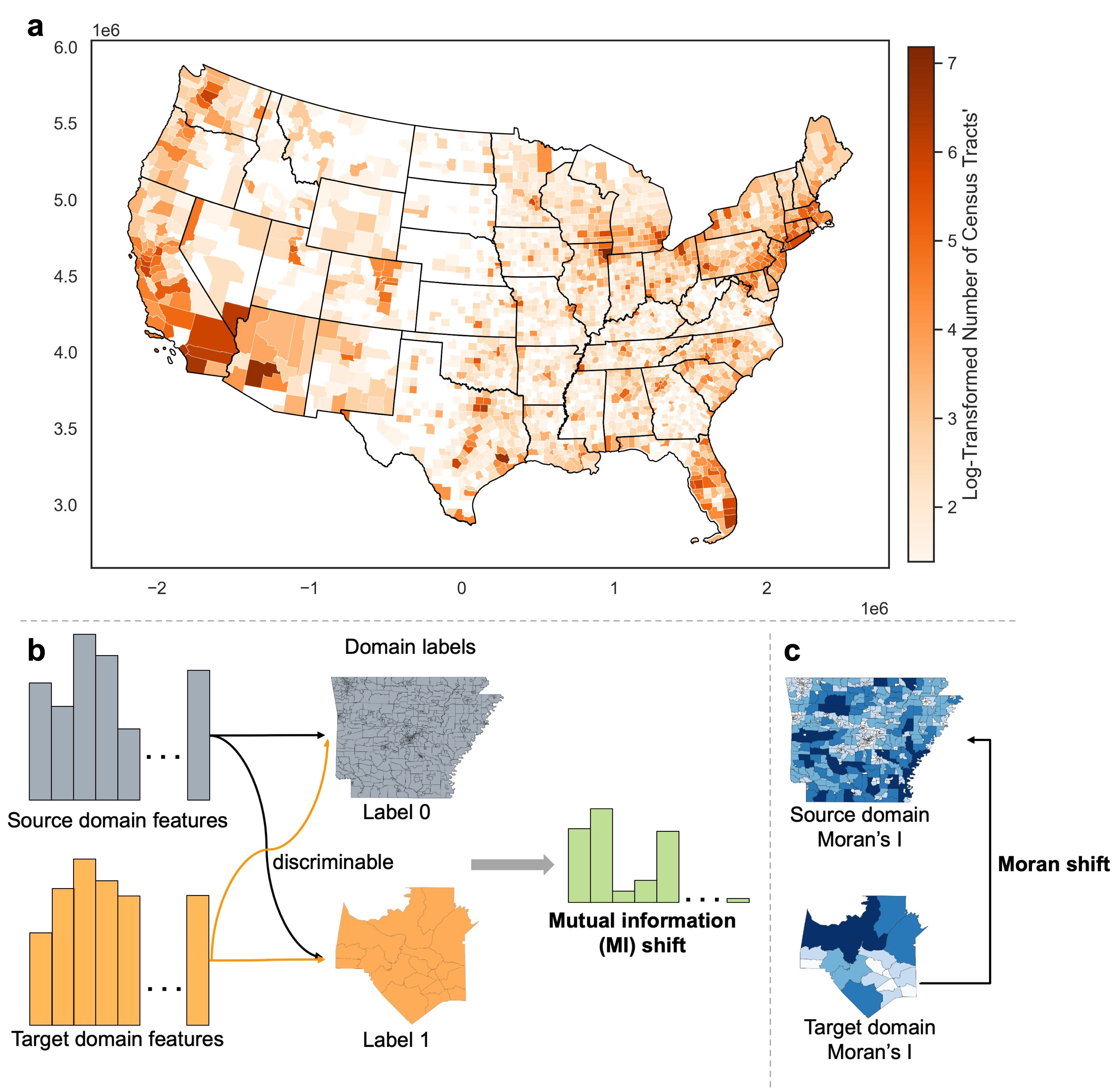}
    \caption{\textbf{a.} The county occupation of the used census tract-level commuting OD flow dataset in the United States. The color saturation denotes the log-transformed census tract number. \textbf{b.} The mutual information (MI) shift between input features of the source domain and input features of the target domain. \textbf{c.} The proposed Moran spatial shift between Moran's I index of the target domain ($I_{tar}$) and that of the source domain $I_{src}$, i.e., $I_{tar} - I_{src}$. }
    \label{fig:framework}
\end{figure}

In this study, we employed an open large-scale OD mobility flow generation benchmark dataset\footnote{\url{https://github.com/tsinghua-fib-lab/CommutingODGen-Dataset}}, \textit{CommutingODGen} \citep{rong2025large}, which provides both input geographic features and output OD flows, to investigate the geospatial transferability of a deep learning–based human mobility generation model. The rationale for using this dataset is twofold: 
\begin{enumerate}
    \item First, the OD flows span a wide range of geographic areas, covering 2,265 counties across 48 states in the United States\footnote{Although the dataset includes 3233 counties from 50 states and Washington D.C. of the United States, we excluded the states of Alaska and Hawaii as well as Washington D.C., and kept 48 states in the Contiguous U.S. for our data analysis.}, as shown in \Cref{fig:framework} (a). This broad spatial coverage enables the systematic cross-region evaluation. Notably, the OD flows are restricted to within-county interactions and mainly depict commuting-related spatial interactions among census tracts. 
    \item  Second, the dataset includes both input geographic features and corresponding OD flows for the year 2018. Specifically, the input geographic features have 97 demographic features and 36 categories of points of interest (POIs), which provide a comprehensive representation of the socioeconomic and built-environment characteristics underlying commuting mobility flows. As these geographic features are well aligned with each census tract within a county, additional alignment effort between input and output is unnecessary, thereby facilitating reproducible analysis. 
\end{enumerate}

% Census tract-level commuting OD flow data within counties
% \subsection{Overall workflow}

\subsection{Quantification of geographic domain shift}
\label{subsec:shift_metrics}
Unlike previous studies, which typically assumed that the training geographic area (i.e., the source geographic domain) is inherently different from the transferring geographic area (i.e., the target geographic domain) when assessing model transferability, we hypothesize that the actual difference between the source and target geographic domains varies across domain pairs. Accordingly, we propose to explicitly quantify geographic domain shift based on the intrinsic characteristics of geographic features and spatial structure used to train a human mobility generation model. Specifically, we introduce \textit{mutual information (MI) shift} to measure differences in the statistical distributions of geographic features between source and target domains, following classical formulations of covariate shift \citep{blitzer2007biographies}. In addition, to account for changes in spatial structure that may not be captured by purely distributional measures, we further propose a \textit{Moran spatial shift}, derived from the spatial autocorrelation metric, Moran's I \citep{moran1950notes}. Together, these two metrics characterize complementary aspects of geographic domain shift: MI shift captures feature-level (attributes) distributional discrepancies, while Moran shift reflects differences in the spatial dependence of features. 

Notably, the quantification of geographic domain shift depends solely on the underlying geographic features, such as the 97 demographic variables and 36 POI categories used in this study, and is independent of any specific human mobility generation model. This model-agnostic design enables the proposed metrics to be applied prior to model training and to support quantitative assessments of transferability across geographic domains.

\begin{definition}[Mutual information shift]
 It measures the extent to which the N-dimensional geographic features can reduce uncertainty in distinguishing between the source and target geographic domains.
\end{definition}

As illustrated in \Cref{fig:framework} (b), given the $i$th dimension of N-dimensional geographic features $X_{i}$ from both source distribution ($p_{src} (X_{i})$) and target distribution ($p_{tar} (X_{i})$) and domain labels,
\begin{equation}
    Y = \begin{cases}
    0,  & \mbox{if $X_{i}$}\mbox{ $\sim p_{src} (X_{i})$} \\
    1, & \mbox{if $X_{i}$}\mbox{ $\sim p_{tar} (X_{i})$}
\end{cases}
\end{equation}
The mutual information shift score for the $i$th dimension of features is calculated as follows~\citep{cover1999elements}, 
\begin{equation}
    MI_{i}(X_{i};Y) = \sum_{x \in X_{i}}\sum_{y \in Y}p(x,y)\log\frac{p(x,y)}{p(x)p(y)}
\end{equation}

As a consequence, the overall MI shift based on N-dimensional geographic features is scored by averaging the sum of the N-dimensional MI shift scores,
\begin{equation}
    MI(X;Y) = \frac{\sum_{i=1}^N MI_{i}}{N}
\end{equation}

The MI shift score essentially measures the dependence of covariates on two geographic domains. A higher MI shift score means that the covariate is more domain-dependent (i.e., knowing the values of features $X$ gives a lot of information about which geographic domain the data come from), indicating a larger shift between the source and target domains.

\begin{definition}[Moran spatial shift]
    As shown in \Cref{fig:framework} (c), it measures the change in the spatial dependence of geographic features between the source and target domains (i.e., changes in spatial autocorrelation patterns), reflecting how features are more spatially clustered or dispersed from the source to the target domain.
\end{definition}
The intuition for the Moran shift metrics is that existing geographic similarity metrics, which are mainly derived from classical distribution distances, such as cosine similarity and Euclidean distance of geographic features, rarely account for the spatial structures of geographic features~\citep{zhao2025multivariate}, while most of the geographic phenomena, including human mobility flows \citep{boucherie2025decoupling,zhao2025predicting}, hold significant spatial dependence and spatial heterogeneity \citep{goodchild2021replication}. Given N-dimensional geographic features of the source and target domains, the corresponding multivariate Moran's I indices \citep{lin2023comparison}, $I_{src}$ and $I_{tar}$, are calculated, respectively. Specifically, the standard formulation of a multivariate Moran's I is represented as,
\begin{equation}
    I = \frac{\sum_{i=1}^M\sum_{j=1}^M w_{ij}Z_{i}^\top Z_{j}}{\sum_{i=1}^MZ_{i}^\top Z_{i}}
\end{equation}
Where $M$ is the total number of geographic areas, $X_{i} = (x_{i1},...,x_{iN})$ represents N-dimensional geographic features at the $i$th area, $Z$ is a column-wise centered and standardized matrix for the original geographic feature matrix $X$, and $w_{ij}$ is a spatial weight between $i$th and $j$th areas, which fulfills $\sum_{j=1}^M w_{ij} = 1$ for each geographic area. 

Hence, the Moran spatial shift from the source to the target geographic domain is calculated as:
\begin{equation}
\label{eq:moran}
    Moran_{src\rightarrow tar} = I_{tar} - I_{src}
\end{equation}

Given this formula, the derived Moran shift score holds the sign: when it is $>0$, it indicates that the covariates in the target geographic domain are more spatially clustered than those in the source domain, and when it is $<0$, it indicates that the covariates in the target domain are more spatially dispersed. And Moran shift = 0 indicates the same overall strength of spatial autocorrelation (clustered, random, or dispersed) between source and target domains, but it does not necessarily show identical spatial patterns between source and target domains in terms of location of clusters and data values.

\subsection{Training of human mobility generation model}
In data-driven human mobility generation, the choice of training and testing data is critical for revealing cross-region model performance, which in turn substantially influences the bias and fairness of the synthesized mobility data \citep{schlosser2021biases, nijs2025data, liu2025generating}. This motivates us to examine the relationships between intrinsic data characteristics and model transferability of human mobility generation models, beyond the extensively studied improvements achieved through model architecture design.

To this end, we chose a popular human mobility generation model, DeepGravity \citep{simini2021deep}, as the generator of census tract-level commuting OD flows within each of 2265 counties. DeepGravity is inspired by the classical human mobility gravity mechanism and formulates the mobility flow generation process as a multinomial logistic regression problem, conditioned on the geographic features of an origin, its distances to potential destinations, and the known total outflow. Its core architecture consists of a feed-forward neural network with 15 hidden blocks, each of which is composed of a linear layer and a LeakyReLU activation layer\footnote{\url{https://github.com/scikit-mobility/DeepGravity}}. Owing to its strong performance and widespread adoption, DeepGravity has become a popular baseline for human mobility flow generation \citep{rong2025large, wang2025deep, xu2025predicting}.

Following the \textit{leave-one-city-out} strategy performed by \citet{simini2021deep}, we adopted a \textit{leave-one-state-out} experimental design to evaluate geospatial transferability of the DeepGravity model. Specifically, for each experiment, data from a single state were used for training and validation, while data from all remaining states were used for testing transfer performance. Within the selected training state, counties were used as the basic geographic units to preserve spatial coherence: 70\% of counties were randomly assigned to the training set and the remaining 30\% to the validation set for hyperparameter tuning. As a result, 48 separate models were trained, one for each state\footnote{Washington, D.C. was excluded due to insufficient data to support a 7:3 train–validation split.}. Given the large scale of the dataset, the batch size for training was set to 512 and the learning rate to $3\times10^{-4}$. Models were trained for up to 100 epochs, with early stopping applied when the validation loss increased for more than 10 consecutive epochs, in order to mitigate overfitting.

To verify the representativeness of DeepGravity in generating human mobility flows and examine the robustness of the associations between the proposed geographic domain shift metrics and the geospatial transferability of a human mobility generation model, we further trained two classical machine learning-based mobility generation models, boosting regression tree (GBRT) \citep{robinson2018machine} and random forest (RF) \citep{pourebrahim2019trip}, and one graph neural network-based mobility generation model, geo-contextual multitask embedding learning (GMEL) \citep{liu2020learning}. While the parameters of these models follow the same settings as reported in \citep{rong2025large}, we solely changed the \textit{max\_depth} of GBRT from None to 3 to speed up the model training.
\subsection{Evaluation of geospatial transferability}
The evaluation of geospatial transferability is to compare the generated census tract-level OD flows $T_{ij}$ with the ground truth commuting flows $\hat{T}_{ij}$ within each target domain testing county. Two metrics were used to evaluate the geospatial transferability of the model trained in a source state domain: the common part of commuting (CPC) and the root mean squared error (RMSE), as calculated by \Cref{eq:cpc} and \Cref{eq:rmse}.
\begin{equation}
\label{eq:cpc}
    CPC= \frac{2\sum_{i,j}min(T_{ij}, \hat{T}_{ij})}{\sum_{i,j}T_{ij} + \sum_{i,j}\hat{T}_{ij}}
\end{equation}
\begin{equation}
\label{eq:rmse}
    RMSE = \sqrt{\frac{1}{N}\sum_{ij}(T_{ij}-\hat{T}_{ij})^2}
\end{equation}

In the context of OD flow generation, the CPC metric measures the overlap between the observed and generated flows, that is, the smaller of the two values. Then it sums the overlap across all OD pairs and normalizes it by total flow volume. It captures the overall spatial structure similarity between the observed and generated flows. A high CPC means the model captures where commuters actually go from origins, while a low CPC means that the model misses important flow patterns and allocates flows to incorrect destinations. In terms of RMSE, it measures the average squared difference between the observed and generated flows, that is, the average magnitude of prediction error. It mainly reflects numerical accuracy more than the structural overlap of mobility flows that are depicted by the CPC metric.
% \begin{itemize}
%     \item Fine-grained county-level transferability
%     \item Aggregated state-level transferability
% \end{itemize}
% \subsubsection{Evaluation metrics}

\subsection{Linear mixed-effects regression model}
\label{subsec:regression}
Furthermore, we want to explore the associations between proposed geographic domain shift metrics and the geospatial transferability of the OD flow mobility generation model. In our study, the geospatial transferability of DeepGravity and the other three models, measured by CPC and RMSE, is typically regarded as a response to the geographic domain shift (measured by MI shift and Moran shift). Because each county was evaluated repeatedly under different source training states, the resulting CPC and RMSE observations were not independent. To account for this repeated-measure structure and unobserved county-level heterogeneity, we included a county-specific random effect. Source training states were modeled as fixed effects because they represent a finite set of experimental conditions under which geospatial transferability was evaluated. Accordingly, a linear mixed-effects regression model was chosen to model the relationship between model transferability and geographic domain shift scores, including MI shift and Moran shift, which is formulated as:
\begin{equation}
    y_{c, s} = \beta_{0} + \beta_{1}MI_{c, s} + \beta_{2}Moran_{c, s} + \gamma_{s} +  \delta_{c} +  \epsilon_{c, s}
\end{equation}
Where $c$ is target county domain and $s$ is a source training state domain, $y_{c, s}$ denotes a transferability metric (CPC or RMSE) in $c$ under the condition of $s$, $\beta_{0}$ is a global intercept, and $\beta_{1}$ and $\beta_{2}$ represent the population-average MI shift effect and Moran shift effect. Moreover, $\gamma_{s}$ is the fixed effect of training state (dummy variable), $\delta_{c}$ is a county-specific random intercept (which draws from $N(0,\sigma_s^2$), accounting for unobserved county-level heterogeneity or variation), and $\epsilon_{c, s}$ denotes the residual error. Notably, since the statistical distribution of RMSE is right-skewed and the value range of CPC is [0,1], we performed a log transformation of the RMSE and a logit transformation of CPC for the regression analysis. The implementation of the linear mixed-effects model (MixedLM) was done using the \textit{statsmodels} library\footnote{\url{https://www.statsmodels.org/stable/examples/notebooks/generated/mixed_lm_example.html}} in Python. 

\section{Results}

\subsection{Predictive performance of human mobility generation models}
The predictive performance of trained mobility generation models is evaluated on the unseen regions from the source training state. As reported in \Cref{tab:pred_eval}, DeepGravity outperforms the other three human mobility generation models, achieving the highest CPC and the lowest RMSE. Notably, the high standard deviations of RMSE values from all four models indicate that there are extreme cases in the predictions of numerical values of OD flows. Given such a comparative performance, we chose the DeepGravity model as the primary focus in our research.

\begin{table}[H]
\tbl{The predictive performance of different mobility generation models}
{
\begin{tabular}{@{}lcccc@{}}
\toprule
\multicolumn{1}{l}{\multirow{2}{*}{\textbf{Models}}} & \multicolumn{2}{c}{\textbf{CPC}} & \multicolumn{2}{c}{\textbf{RMSE}} \\ \cmidrule(l){2-5} 
\multicolumn{1}{l}{} & \multicolumn{1}{c}{Mean} & \multicolumn{1}{c}{Std Dev} & \multicolumn{1}{c}{Mean} & \multicolumn{1}{c}{Std Dev} \\ \midrule
DeepGravity & 0.603 & 0.101 & 82.533 & 61.706 \\
RF & 0.558 & 0.115 & 90.061 & 68.789 \\
GBRT & 0.525 & 0.117 & 93.515 & 70.256 \\
GMEL & 0.464 & 0.164 & 146.94 & 97.759 \\ \bottomrule
\end{tabular}
}

\label{tab:pred_eval}
\end{table}
\subsection{Characteristics of geospatial transferability of mobility generation}
\label{subsec:transferability_eval}

\begin{figure}[!ht]
    \centering
    \includegraphics[width=1\textwidth]{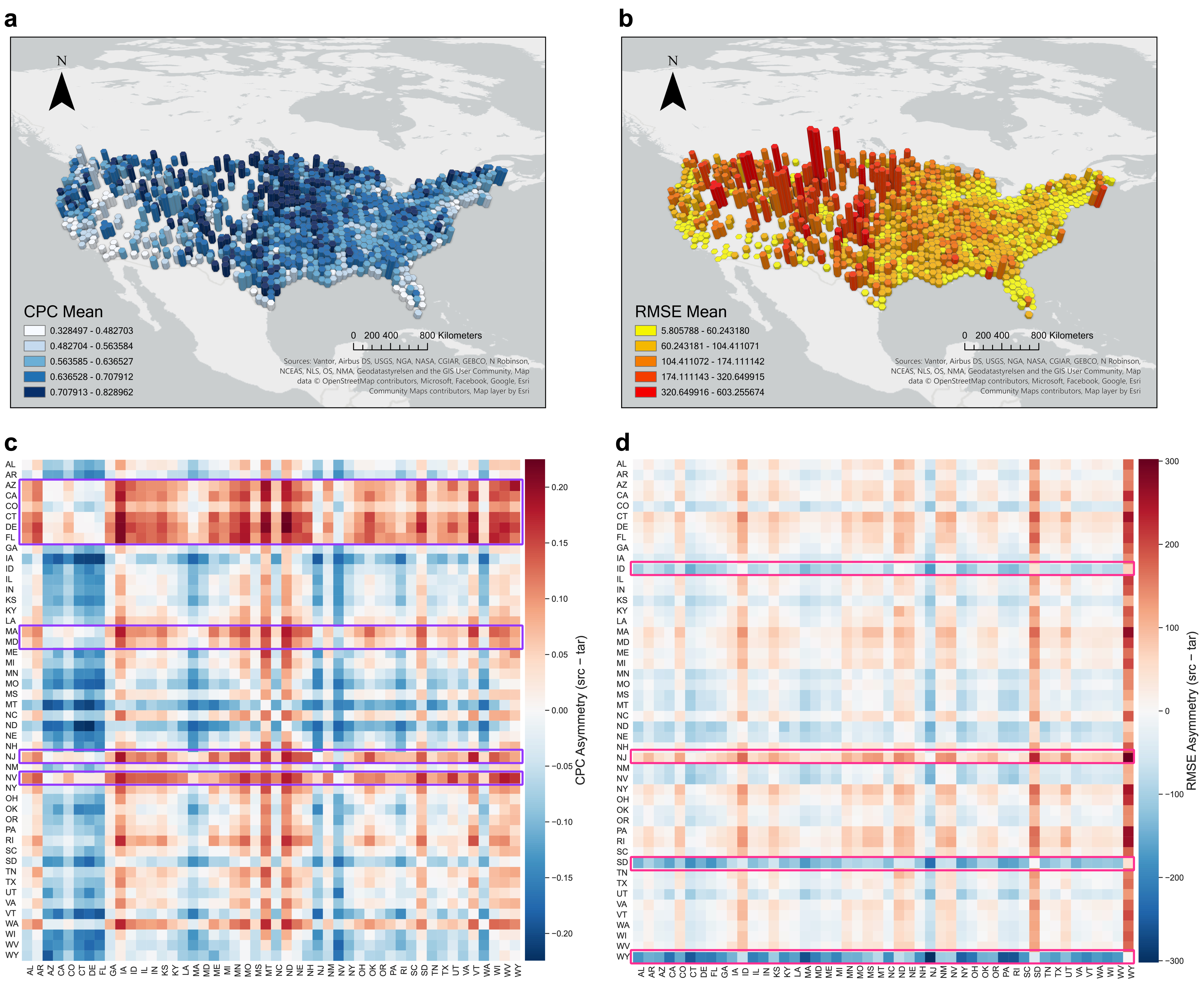}
    \caption{The hexagon cube maps of OD flow generation transfer performance, \textbf{a.} CPC and \textbf{b.} RMSE, which were aggregated with hexagon cells across the 2265 counties from 48 states, where the counties of a state construct the source domain for training, and the remaining counties are transferred to. The color saturation of a hexagon cube represents the mean values of CPC or RMSE achieved in target counties within the hexagon area, while the height of the hexagon cube denotes the standard deviation of CPC or RMSE achieved by DeepGravity models trained with different source states in the same hexagon area. Notably, the height in \textbf{a.} is scaled up to 5000 times the CPC std., and the height in \textbf{b.} is scaled up to 10 times the RMSE std. The bidirectional evaluation of state pairs, one of which acts as a source state domain for training and the other one is a target domain for transfer, shows the asymmetry of geospatial transferability regarding \textbf{c.} CPC and \textbf{d.} RMSE. Each row represents the geospatial transferability from a source state to the other states.}
    \label{fig:performance}
\end{figure}

\noindent\textbf{Spatial heterogeneity of  transferability.} \Cref{fig:performance} (a) and (b) illustrate the transfer performance of the DeepGravity model trained with different source states, measured by CPC and RMSE, respectively. The achieved overall CPC and RMSE by the DeepGravity model hold significant spatial heterogeneity, as shown in the varying saturation of colors in the figures. It is worth noting that CPC generally captures global mobility flow patterns, whereas RMSE emphasizes absolute flow magnitudes at individual spatial units and is more sensitive to outliers. Therefore, the two metrics show distinctive spatial patterns.

Specifically, DeepGravity achieved a high CPC (over 0.70) mainly in the Midwestern states of U.S., i.e., dark blue areas in \Cref{fig:performance} (a). In contrast, in those areas along the east and east coasts, the mean CPC values of DeepGravity models trained with all of the individual source states are much lower, e.g., in the range of [0.33, 0.48] for light blue areas, which indicates a more complex human mobility flow structure that cannot be easily learned from other states. Meanwhile, those tall hexagon cubes indicate the larger standard deviation of CPC by different individual source state-trained models, indicating inconsistent performance of the same deep learning architecture (i.e., DeepGravity in this study) across source training states. Geographically, these areas of high CPC standard deviation are also areas of high mean values of CPC.

Regarding the RMSE-based transfer performance, the generated OD flows by  DeepGravity models in the Western and Midwestern U.S. states held substantial errors, and the change of the individual source states for training resulted in considerable changes in RMSE, as indicated by the high saturation of orange and tall heights of hexagon cubes. Furthermore, it is visually clear that the overall RMSE gradually decreases from the Midwestern to the Eastern states, indicating increased transfer performance, except for several areas in Maine and Florida. The distinct patterns captured by CPC and RMSE metrics reflect the different nature of the measurement (i.e., global mobility structure v.s. absolute flow magnitude).

\noindent\textbf{Asymmetry in geospatial transferability.} For each DeepGravity model trained on a single source state, we aggregated CPC and RMSE values across target testing counties according to their corresponding states, thereby deriving state-to-state transfer performance. To examine whether a pair of states can transfer symmetrically between their respective domains, we constructed an asymmetry metric defined as the difference between transfer performance from the source state to the target state and that from the target state to the source state. Accordingly, the difference is directional and signed, where values greater than zero indicate that the DeepGravity model trained on the source state exhibits higher geospatial transferability to the target state than the model trained on the target state when evaluated on the source state.

\Cref{fig:performance} (c) and (d) present the CPC and RMSE asymmetry matrices, respectively. Specifically, the CPC asymmetry of DeepGravity ranges from 0 to 0.2, with models trained on Arizona (AZ), California (CA), Colorado (CO), Connecticut (CT), Delaware (DE), Florida (FL), New Jersey (NJ), and Nevada (NV) generally achieving lower CPC when transferred to other states than models trained elsewhere, indicating their heterogeneous OD flow data distributions. In terms of RMSE, models trained on data from Idaho (ID), South Dakota (SD), and Wyoming (WY) tended to underestimate absolute flows in other states, whereas models trained on New Jersey (NJ) typically overestimated absolute origin–destination flows when applied outside the source state.

% \subsection{Complementary characterization of geographic domain shift via Moran’s I shift}
\subsection{Distribution of geographic domain shift}
\label{subsec:geoshift_distribution}
The geographic domain shift from a source state domain (for training) to the target county domain (for evaluation) is quantified based on MI shift and Moran shift as introduced in \Cref{subsec:shift_metrics}. Notably, since the used benchmark dataset only provides a county-level multivariate distance matrix and misses the raw geographic information of census tracts for OD flows, we can only calculate the county-level Moran's I index, and then compute the average Moran's I of multiple counties in a state to denote the source state's Moran's I. After that, the source-target Moran spatial shifts are calculated according to Equation \eqref{eq:moran}.

 \begin{figure}[!ht]
    \centering
    \includegraphics[width=0.95\textwidth]{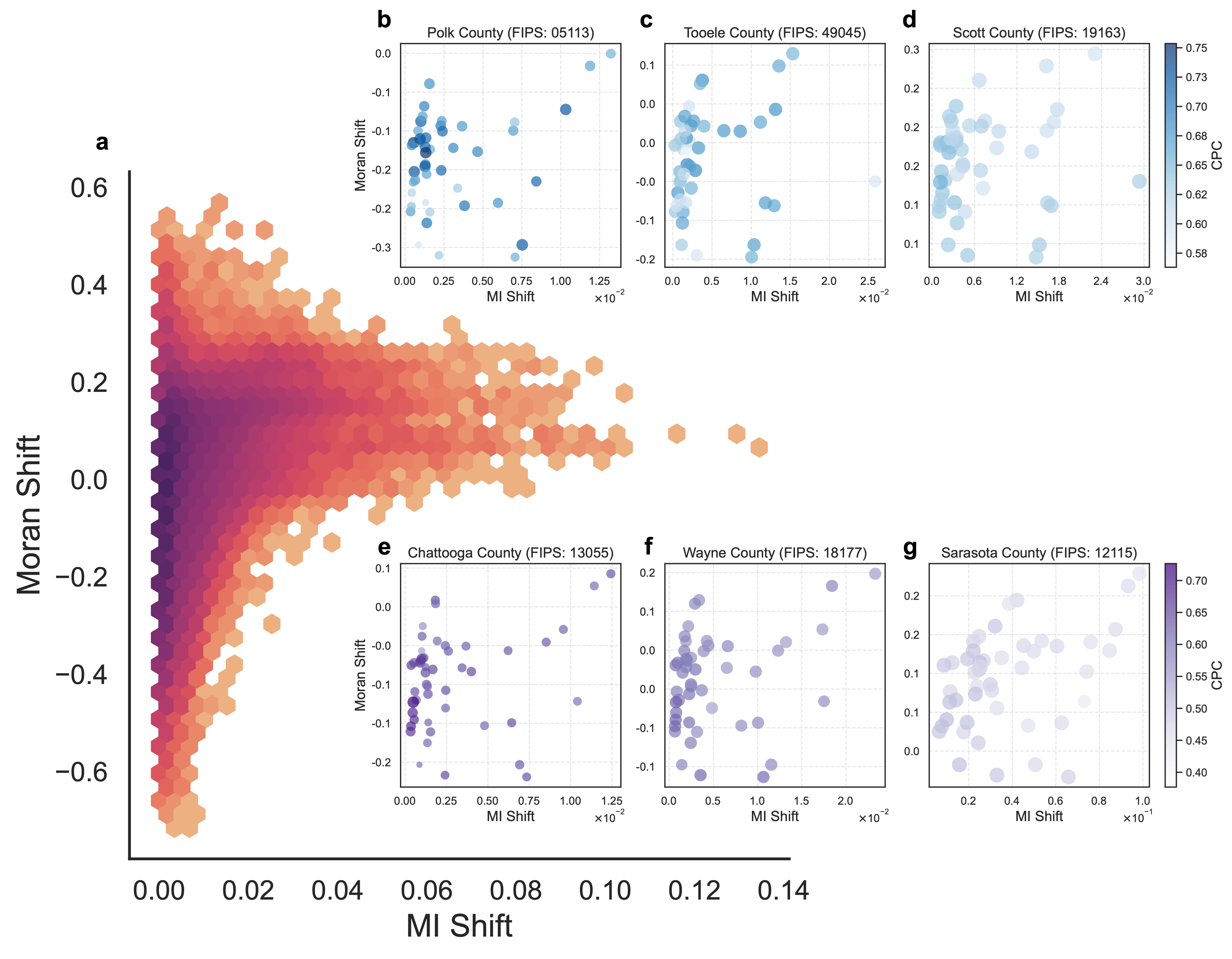}
    \caption{\textbf{a.} The joint distribution of mutual information (MI) shift (x-axis) and Moran's I spatial shift (y-axis). \textbf{b-d.} The MI-Moran's I shift distribution of different reference states in three counties, where the MI shifts remain at the 25\%, 50\%, and 75\% quantiles in ascending order, respectively, and \textbf{e-g.} The distribution in three counties whose Moran's I index shifts remain at the 25\%, 50\%, and 75\% quantiles, respectively. The saturation of color denotes the CPC transferability, with darker indicating a higher CPC. The size of dots represents the RMSE transferability, with the larger size meaning a larger RMSE.}
    \label{fig:joint_distribution}
\end{figure}

\Cref{fig:joint_distribution} (a) visualizes their joint distribution of MI shift and Moran shift. While the MI shift of domain pairs ranges from 0.00 to 0.14 generally, the majority stay in the range of [0.00, 0.06]. Regarding Moran shift, it changes between -0.75 and 0.6 approximately, mostly staying in the range of [-0.4, 0.2]. Notably, a positive sign of Moran shift indicates that the spatial structures of geographic features in the target county domain are typically more clustered than the spatial structures shown in the source state domain, while a negative sign indicates more dispersed spatial structures. Moreover, the shape of the joint distribution of MI shift and Moran shift reveals that Moran shift varies even though MI shift is very minor, e.g., around 0.00, implying the non-collinear relationship of the proposed Moran shift with MI shift. The two metrics capture different aspects of the geographic domain shift.

Additionally, we depicted the joint distributions of three counties based on the quantiles of MI shifts (\Cref{fig:joint_distribution} b-d) and the quantiles of Moran shifts (\Cref{fig:joint_distribution} e-g), respectively, to illustrate the details of the geospatial transferability evaluations at a county. In each scatter figure, 47 out of 48 source states (excluding the state containing the transferred state) result in 47 Moran shifts and MI shifts from their source domain to the same target county domain, and the corresponding transferability of the trained DeepGravity models. Generally, the dots in each figure are dispersed along both MI shift and Moran shift axes, indicating that the geographic domain varies with the source state domains, given the same target county domain. Among the three counties with an increase of MI shift (\Cref{fig:joint_distribution} b-d), from Polk County in Arkansas, Tooele County in Utah, to Scott County in Iowa, their Moran shifts generally increase, and their scatter ranges are from [-0.3, 0.0] to [-0.2, 0.1] and [0.1, 0.3]. 

In terms of the Moran shifts, the 25\% quantile is Chattooga County in Georgia, the 50\% quantile is Wayne County in Indiana, and the 75\% quantile is Sarasota County in Florida. In particular, the corresponding MI shift of Wayne County (\Cref{fig:joint_distribution}f, at the 50\% quantile Moran shift level) span over the largest range, [0.0, 2.5]. Additionally, the color saturation and size of dots across target county domains demonstrate that the DeepGravity models typically perform better under lower MI shift (e.g., \Cref{fig:joint_distribution}b) and Moran shift (e.g., \Cref{fig:joint_distribution}d) conditions. However, within each target county, the variance of transferability caused by the change of source state domains cannot be overlooked, which is a detailed illustration of high cubes in 3D hexagon choropleth maps as shown in \Cref{fig:performance}.

\subsection{Associations of geographic domain shifts with geospatial transferability}

To understand the associations of geographic domain shifts with the geospatial transferability of individual source state-trained models at different target county domains, we fitted the linear mixed-effects regression models (see Methods Section \ref{subsec:regression}) with CPC and RMSE on target counties for the four trained models (i.e, DeepGravity, RF, GBRT, and GMEL), respectively. As reported in \Cref{tab:mixed_effect}, MI shift is negatively associated with CPC, indicating improved transferability under smaller mutual-information shift, whereas Moran shift exhibits a positive association with CPC, suggesting improved geospatial transferability when the spatial autocorrelation pattern in the target domain is stronger than that in the source domain. According to the regression for log-transformed RMSE, we found that MI shift is positively associated with \texttt{log(RMSE)}, indicating increased RMSE, i.e., poorer transfer performance, under larger mutual-information shift. In contrast, Moran shift exhibits a negative association with \texttt{log(RMSE)}, suggesting smaller RMSE, i.e., improved geospatial transferability, with a stronger spatial autocorrelation pattern in the target domain.

% The patterns are consistent with results obtained using CPC as an alternative transferability metric, demonstrating robustness across evaluation measures.
\begin{table}[H]
\tbl{Linear mixed-effects regression results for geospatial transferability responses.}
{
\begin{tabular}{@{}llcccc@{}}
\toprule
\multirow{2}{*}{\textbf{Models}} & \multirow{2}{*}{\textbf{Predictors}} & \multicolumn{2}{c}{\textbf{\begin{tabular}[c]{@{}c@{}}CPC (Logit-transformed)\\ response\end{tabular}}} & \multicolumn{2}{c}{\textbf{\begin{tabular}[c]{@{}c@{}}RMSE (Log-transformed) \\ response\end{tabular}}} \\ \cmidrule(l){3-6} 
 &  & \textbf{Coefficients} & \textbf{p-value} &  \textbf{Coefficients} & \textbf{p-value}\\ \midrule
\multirow{2}{*}{DeepGravity} & MI shift & -0.023 & \textless{}0.001 & 0.010 & \textless{}0.001 \\
 & Moran shift & 0.013 & \textless{}0.001 & -0.007 & \textless{}0.001 \\ \midrule
\multirow{2}{*}{RF} & MI shift & -0.357 & \textless{}0.001 & 0.239 & \textless{}0.001 \\
 & Moran shift & 0.178 & \textless{}0.001 & -0.097 & \textless{}0.001 \\ \midrule
\multirow{2}{*}{GBRT} & MI shift & -0.430 & \textless{}0.001 & 0.287 & \textless{}0.001 \\
 & Moran shift & 0.175 & \textless{}0.001 & -0.098 & \textless{}0.001 \\ \midrule
\multirow{2}{*}{GMEL} & MI shift & -0.213 & \textless{}0.001 & 0.149 & \textless{}0.001 \\
 & Moran shift & 0.071 & \textless{}0.001 & -0.058 & \textless{}0.001 \\ \bottomrule
\end{tabular}
}
\label{tab:mixed_effect}
\end{table}

Since a higher CPC refers to better geospatial transferability but RMSE is the opposite, the regression analysis results for both transferability metrics jointly imply that a higher MI shift (i.e., more distinctive feature distributions) is associated with a poorer transfer performance of a mobility generation model, while a stronger spatial autocorrelation can inform better transferability. Moreover, the signs of associations between the proposed two geographic domain shift metrics and the two transferability metrics are consistent across all the four trained mobility generation models, which suggests generalizable associations independent of specific human mobility generation models.
\begin{figure}[!ht]
    \centering
    \includegraphics[width=1\textwidth]{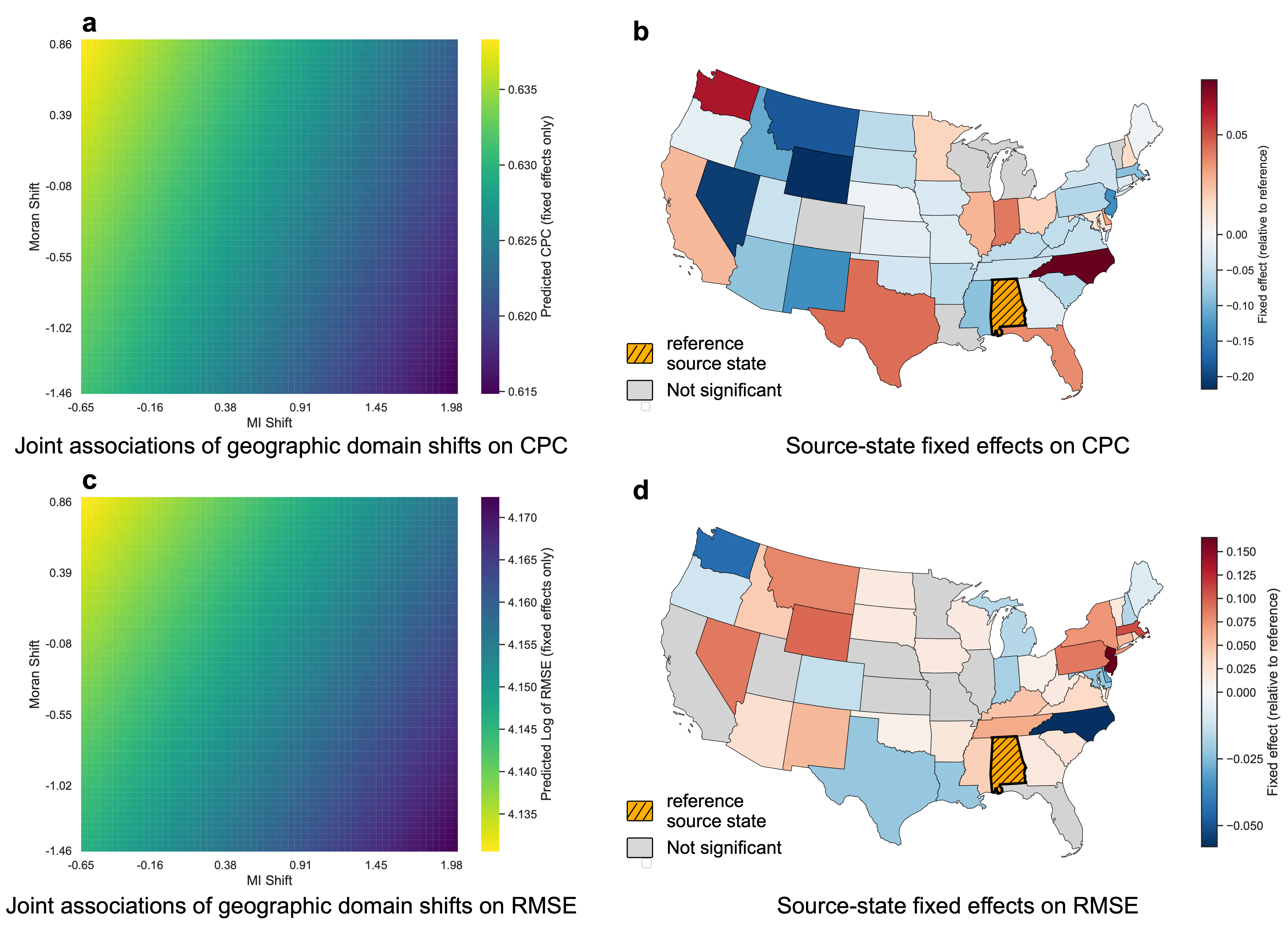}
    \caption{The joint associations of MI shift and Moran shift on CPC in \textbf{(a)} and on RMSE in \textbf{(c)} of fitted linear mixed-effects models based on DeepGravity. The derived fixed effects of source (training) states are mapped in \textbf{(b)} and \textbf{(d)}, respectively.}
    \label{fig:shift4metrics}
\end{figure}

Furthermore, the joint associations of MI shift and Moran shift with geospatial transferability of DeepGravity in fitted models are visualized in \Cref{fig:shift4metrics} (a) and (c) regarding CPC and RMSE, respectively. Moreover, the source-state fixed effects were mapped in \Cref{fig:shift4metrics} (b) and (d), where 6 out of 48 source states did not show statistically significant fixed effects on CPC, while 8 source states were not statistically significant in the regression of log-transformed RMSE.

\section{Discussion}
\subsection{Biased evaluations of geospatial transferability in human mobility generations}
The systematic leave-one-state-out evaluations in \Cref{subsec:transferability_eval} demonstrate that geospatial transferability exhibits substantial spatial heterogeneity and asymmetry across state pairs, regardless of whether CPC or RMSE is used. These findings indicate that the transfer performance of a human mobility generation model is biased by the choice of source-domain training data and the target-domain transfer setting. However, existing human mobility generation studies \citep{rong2023goddag, rong2025large, li2025cross} typically rely on a small number of regions and averaged overall evaluation metrics (e.g., mean CPC and mean RMSE) to compare models and assess cross-region generalization. While this evaluation practice is reasonable and facilitates straightforward performance comparisons across different deep learning–based mobility generation models, it risks obscuring the models’ transferability across geographic settings.

Specifically, changes in the source-domain training data or the target-domain testing region may lead to substantial performance variation, making it uncertain whether an initially well-performing model will maintain its effectiveness across regions. As illustrated in \Cref{fig:performance}, the mean CPC of DeepGravity models trained on different individual source states ranges approximately from 0.58 to 0.65, yielding a difference of about 0.07. Notably, some existing studies report CPC differences between baseline models that are smaller than this range, which might confound whether the observed performance differences arise from model architectures or from the intrinsic characteristics of the datasets.

Additionally, we observe that transfer performance assessed using CPC is not always consistent with that evaluated using RMSE. In other words, these regions with higher CPC are prone to getting larger RMSE. However, this discrepancy is not contradictory, because CPC captures global mobility flow patterns and their overall spatial structure similarity, while RMSE emphasizes absolute flow magnitudes at individual spatial units and is therefore more sensitive to outliers \citep{rong2024interdisciplinary}. Such an inconsistency indicates that geospatial transferability is a concept of multiple dimensions and may be constrained by geographic scales. As a consequence, different evaluation metrics can reflect different aspects of geospatial transferability. Furthermore, it highlights a fundamental challenge in human mobility generation, that is, how to make a trade-off between accurately reproducing global patterns and capturing fine-grained local details \citep{mauro2022generating}.

\subsection{Asymmetric geospatial transferability}
As illustrated in \Cref{fig:performance} (c) and (d), the geospatial transferability, as evaluated by both CPC and RMSE, exhibits substantial asymmetry. Specifically, if an OD flow generation model trained in a source region can generate realistic OD flows in a target region, the converse does not necessarily hold: a model trained in the target region typically does not achieve equivalent transferability to the source region. Although such asymmetry in transferability has been observed in many mobility generation studies \citep{he2020human,rong2023goddag} and other geospatial data generation studies \citep{yu2025exploring}, this study is the first to formally conceptualize it. This asymmetry is likely driven by heterogeneous geographic structures between source and target regions, whereby models trained in the source region may fail to capture the more complex relationships present in the target region.

More importantly, this transfer asymmetry further supports our rationale for introducing a signed metric, such as the Moran shift proposed in this study. Unlike conventional domain shift measures that capture solely distributional differences or similarities, such as mutual information shift, the Moran shift can explicitly represent the direction of transfer from the source region to the target region. Additionally, although classical theories of geographic similarity \citep{zhu2022third} are related to geographic domain shift, most similarity metrics are also directional \citep{yu2025exploring}. This further highlights the significance and timeliness of our proposed Moran shift as a new measure for quantifying geographic domain differences.

\subsection{Feasibility of the proposed geographic domain shift metrics}
Given the varying transfer performance across source-target domain pairs, we hypothesize that they are primarily influenced by the intrinsic data characteristics of datasets encompassing both geographic domains. Accordingly, this study proposes two metrics, mutual information (MI) shift and Moran shift, to capture intrinsic covariate shifts from the source to the target geographic domain. The MI shift is a classical metric for quantifying covariate shift in terms of distributional similarity, whereas the Moran shift is newly introduced to characterize directional changes in the spatial structure of geographic features between the source and target domains.

As demonstrated in \Cref{subsec:geoshift_distribution}, the newly introduced Moran spatial shift captures additional covariate shift information that is not characterized by the classical MI shift. Together with the MI shift, it is further shown to be indicative of model transferability, as indicated by the CPC and RMSE of human mobility generation, based on analyses using linear mixed-effects regression models in \Cref{subsec:regression}. These results, in turn, support our hypothesis that geographic domain shifts contribute to biased evaluations of geospatial transferability. Moreover, the MI shift and Moran shift may complement geographic similarity in explaining the cross-region model transferability for other geospatial tasks \citep{yu2025exploring}.

Moreover, geographic domain shift has received considerable attention in the earth observation and remote sensing communities \citep{kalluri2023geonet,al2025analysing,al2025benchmarking,crasto2025robustness,doerksen2026earthshift}. As their primary geospatial tasks typically involve semantic and instance segmentation of satellite imagery, existing studies mainly focus on shifts in the distribution of geographic feature classes rather than their spatial structures. Therefore, the proposed geographic domain shift metrics remain valuable for providing deeper insights into the intrinsic characteristics of satellite image datasets.

\subsection{Association between training dataset size and transferability}
\label{subsec:data_size}
Since our research adopted the \textit{leave-one-state-out} strategy to repeatedly evaluate the model transferability of mobility generation models on held-out target counties, the training dataset size, i.e., the number of census tracts acting as origins and destinations in our used mobility OD flow benchmark dataset, is inevitably not the same across states. Hence, it may act as a confounding factor in the association analysis of transfer performance with the introduced MI shift and Moran shift, given the common belief in scaling laws in deep learning, where the size of a training dataset fed into neural networks is one of the important drivers \citep{hestness2017deep, cherti2023reproducible, merchant2023scaling}.

\begin{table}[!ht]
\tbl{Ordinary least squares regression results of geospatial transferability with dataset size, mutual-information shift, and Moran shift. Since the mean CPC distribution is bounded to 0 and 1, it was logit-transformed in a linear regression.}
{
\begin{tabular}{@{}llcl@{}}
\toprule
\textbf{Responses} & \textbf{Predictors} & \textbf{Coefficients} & \textbf{p-value} \\ \midrule
\multirow{3}{*}{Mean CPC (Logit-transformed)} & Log-transformed data records & 0.015 & 0.171 \\
 & mean MI shift & -0.026 & 0.018 \\ 
  & mean Moran shift & 0.003 & 0.674 \\ \midrule
\multirow{3}{*}{Mean RMSE} &  Log-transformed data records & 0.006 & 0.496 \\
 &mean MI shift& 0.012 & 0.152 \\ 
  & mean Moran shift& -0.015 & 0.018 \\ \bottomrule
\end{tabular}
}
\label{tab:data_effect}
\end{table}

To examine the association between dataset size and transferability, we averaged the MI shift and Moran shift across held-out target counties, along with the corresponding CPC and RMSE. As a result, the mean MI shift value, the mean Moran shift value, and the census tract number act as the three intrinsic characteristics of a source-state training dataset. We conducted a regression analysis with the ordinary least squares method. As reported in \Cref{tab:data_effect}, only the mean MI shift remains a significant negative association with the mean CPC (logit-transformed), while in the regression of mean RMSE, only the mean Moran shift holds a significant negative association. This indicates that the proposed geographic domain shift metrics may serve an essential role in influencing the overall transferability of a machine learning model.

\subsection{Implications of geographic domain shift for human mobility generation}
% \subsubsection{For dataset curation}
% \subsubsection{For model optimization}
The broader implications of geographic domain shift for data-driven human mobility generation are twofold:
\begin{enumerate}
    \item \textit{Dataset curation.} \Cref{subsec:data_size} reveals a counterintuitive result: the training dataset size is not the primary driver of the transferability of our human mobility generation model, DeepGravity, compared with the MI shift and Moran shift. This finding suggests that dataset curation for human mobility generation should move beyond simply increasing data volume and instead prioritize intrinsic dataset characteristics, such as the proposed geographic domain shift metrics, to achieve higher geospatial transferability \citep{janowicz2025geofm, stewart2025torchgeo}. The finding links to the \textit{dataset diversity} research in the general machine learning theory \citep{mandal2021dataset, aroyo2023dices, zhao2024measuring, nougnanke2025dataset}. While many studies have mentioned geographic bias as one aspect of dataset diversity, the proposed metrics may be able to provide more insights about the dataset diversity from the perspective of spatial distribution.
    \item \textit{Human mobility generation model development.} The linear mixed-effects regression analysis reveals fixed effects of the source training state on transfer performance of the DeepGravity model, consistent with previously reported fixed effects of geographic areas \citep{chen2025addressing}. This finding motivates the explicit incorporation of spatial structure in developing geographically transferable deep learning models for human mobility generation. For example, aggregation of geographic information in convolutional architectures has been refined by accounting for spatial heterogeneity and autocorrelation of geographic phenomena \citep{deng2025geoaggregator, guo2025regiongcn}. Furthermore, as improving transferability commonly involves reducing domain shifts through feature selection and representation learning \citep{pan2009survey, farahani2021brief}, the newly introduced Moran shift metric, together with the MI shift, can be naturally integrated into model optimization to enhance the geospatial transferability of human mobility generation models and other task-specific deep learning models.
\end{enumerate}

\subsection{Limitations and future work}
Despite the systematic investigation of the associations between intrinsic dataset characteristics and the geospatial transferability of human mobility generation models, this study has several limitations that warrant future work. First, our analysis relies on a single mobility flow dataset in one country, which may introduce partial dependence between the derived geographic domain shift metrics and the transferability measures (i.e., CPC and RMSE). Although we employ a linear mixed-effects regression model to account for fixed effects of source training states and random effects of target counties, more diverse mobility datasets (e.g., different geographic scales, mobility types, and countries) and more themes of geospatial data generation tasks (e.g., population synthesis) are needed in future work~\citep{moska2025obsr} to derive more independent estimates of geographic domain shifts and transferability and to more directly examine their relationships. Second, the quantitative associations between geographic domain shift and geospatial transferability should be further validated using other human mobility generation models or algorithms, to assess whether the indicative power of the proposed MI shift and Moran shift generalizes across different modeling approaches. Moreover, additional metrics should be developed in the future to quantify a broader range of spatial structure changes (e.g., polycentric versus monocentric patterns) beyond the global Moran shift introduced in this study, which solely focuses on the overall spatial autocorrelation of geographic features.

\section{Conclusion}
This study systematically investigates the geospatial transferability of representative human mobility generation models using the large-scale \textit{CommutingODGen} dataset, which covers census-tract–level origin–destination flows across 2,265 counties in 48 U.S. states. Motivated by the role of intrinsic dataset characteristics in shaping transfer performance across source and target domains, we propose two geographic domain shift metrics, mutual information shift and Moran shift, to quantify domain shifts within a dataset. We further employ a linear mixed-effects regression model to disentangle the associations between intrinsic dataset characteristics and the model transferability. The results reveal heterogeneous and asymmetric patterns of transfer performance between the source and target domains, and demonstrate the complementary roles of mutual information shift and Moran spatial shift in characterizing geographic domain shift. Furthermore, the regression analysis shows significant associations between the proposed domain shift metrics and the model transferability, demonstrating the feasibility of these metrics for assessing the intrinsic characteristic differences between training and testing geographic datasets. Therefore, these metrics have considerable potential to guide the selection of geographic training datasets for developing more geographically transferable human mobility generation models and other GeoAI models.

\section*{Data and code availability statement}
The information about the data and code of this study is publicly available at GitHub: \url{https://github.com/GeoDS/GeoDomainShift-Mobility}.

\section*{Acknowledgements}
Zhiyong Zhou sincerely acknowledges the Postdoc.Mobility Fellowship (No. 235381) funded by the Swiss National Science Foundation in supporting this research. 

\section*{Disclosure statement}
No potential conflict of interest was reported by the authors.

\section*{Notes on contributors}
\noindent \textbf{Zhiyong ZHOU} is an Assistant Professor at the School of Earth Sciences, Zhejiang University and was a postdoc at the Department of Geography, University of Wisconsin-Madison. E-mail: zhiyong.zhou@zju.edu.cn. His research interests include cartography, geospatial data science and GeoAI.

\noindent \textbf{Song GAO} is a Professor of Geographic Information Science at the Department of Geography, University of Wisconsin-Madison, WI, 53706. E-mail: song.gao@wisc.edu. His research interests include geospatial data science, GeoAI, social sensing, spatial networks and human mobility.

\noindent \textbf{Qianheng ZHANG} is a PhD student at the Department of Geography, University of Wisconsin-Madison, WI 53706. E-mail: qianheng.zhang@wisc.edu. His research interests include ecology, geospatial data science and GeoAI.

\noindent \textbf{Feng ZHANG} is a Professor at the School of Earth Sciences, Zhejiang University. E-mail: zfcarnation@zju.edu.cn. Her research focuses on spatio-temporal foundation model, urban intelligent computing, and intelligent remote sensing and disaster applications.

\noindent \textbf{Zhenhong DU} is a Professor at the School of Earth Sciences, Zhejiang University. He is the Director of Zhejiang Key Laboratory of Geographic Information Science. Email: duzhenhong@zju.edu.cn. His research interests include AI for Earth science, Earth system modeling, and spatio-temporal big data.

%\newpage
\bibliographystyle{tfv}
\bibliography{geobias}

@article{yang2026transferable,
  title={Transferable human mobility network reconstruction with neuroGravity},
  author={Yang, Jinming and Huang, Shaoyu and Huang, Zongyuan and Jin, Yaohui and Yang, Xiaokang and Gonz{\'a}lez, Marta C and Xu, Yanyan},
  journal={Nature Computational Science},
  volume={6},
  number={6},
  pages={630--641},
  year={2026},
  publisher={Nature Publishing Group}
}

@article{shi2025defining,
  title={Defining concept drift and its variants in research data management: A scientometric case study on geographic information science},
  author={Shi, Meilin and Janowicz, Krzysztof and Liu, Zilong and Karimi, Mina and Majic, Ivan and Fortacz, Alexandra},
  journal={Transactions in GIS},
  volume={29},
  number={3},
  pages={e70058},
  year={2025},
  publisher={Wiley Online Library}
}

@article{zhu2026gravity,
  title={Gravity-informed deep flow inference for spatial evolution modeling in panel data},
  author={Zhu, Di and Ma, Zhongfu},
  journal={International Journal of Geographical Information Science},
  volume={40},
  number={4},
  pages={918--946},
  year={2026},
  publisher={Taylor \& Francis}
}

@article{doerksen2026earthshift,
  title={EarthShift: a benchmark for measuring robustness to real-world distribution shifts in Earth observation},
  author={Doerksen, Kelsey and Kerner, Hannah},
  journal={arXiv preprint arXiv:2605.29330},
  year={2026}
}

@book{gao2023handbook,
  title={Handbook of geospatial artificial intelligence},
  author={Gao, Song and Hu, Yingjie and Li, Wenwen},
  year={2023},
  publisher={CRC Press}
}

@book{cover1999elements,
  title={Elements of information theory},
  author={Cover, Thomas M},
  year={1999},
  publisher={John Wiley \& Sons}
}

@article{noi2022assessing,
  title={Assessing COVID-induced changes in spatiotemporal structure of mobility in the United States in 2020: a multi-source analytical framework},
  author={Noi, Evgeny and Rudolph, Alexander and Dodge, Somayeh},
  journal={International Journal of Geographical Information Science},
  volume={36},
  number={3},
  pages={585--616},
  year={2022},
  publisher={Taylor \& Francis}
}

@article{xu2016another,
  title={Another Tale of Two Cities: Understanding Human Activity Space Using Actively Tracked Cellphone Location Data},
  author={Xu, Yang and Shaw, Shih-Lung and Zhao, Ziliang and Yin, Ling and Lu, Feng and Chen, Jie and Fang, Zhixiang and Li, Qingquan},
  journal={Annals of the American Association of Geographers},
  volume={106},
  number={2},
  pages={489--502},
  year={2016}
}

@article{xu2021tourism,
  title={Tourism geography through the lens of time use: A computational framework using fine-grained mobile phone data},
  author={Xu, Yang and Li, Jingyan and Xue, Jiaying and Park, Sangwon and Li, Qingquan},
  journal={Annals of the American Association of Geographers},
  volume={111},
  number={5},
  pages={1420--1444},
  year={2021},
  publisher={Taylor \& Francis}
}

@article{xu2018human,
  title={Human mobility and socioeconomic status: Analysis of Singapore and Boston},
  author={Xu, Yang and Belyi, Alexander and Bojic, Iva and Ratti, Carlo},
  journal={Computers, Environment and Urban Systems},
  volume={72},
  pages={51--67},
  year={2018},
  publisher={Elsevier}
}

@article{huang2022staying,
  title={Staying at home is a privilege: Evidence from fine-grained mobile phone location data in the United States during the COVID-19 pandemic},
  author={Huang, Xiao and Lu, Junyu and Gao, Song and Wang, Sicheng and Liu, Zhewei and Wei, Hanxue},
  journal={Annals of the American Association of Geographers},
  volume={112},
  number={1},
  pages={286--305},
  year={2022},
  publisher={Taylor \& Francis}
}

@article{hou2021intracounty,
  title={Intracounty modeling of COVID-19 infection with human mobility: Assessing spatial heterogeneity with business traffic, age, and race},
  author={Hou, Xiao and Gao, Song and Li, Qin and Kang, Yuhao and Chen, Nan and Chen, Kaiping and Rao, Jinmeng and Ellenberg, Jordan S and Patz, Jonathan A},
  journal={Proceedings of the National Academy of Sciences},
  volume={118},
  number={24},
  pages={e2020524118},
  year={2021},
  publisher={National Academy of Sciences}
}

@article{hou2025transferred,
  title={Transferred bias uncovers the balance between the development of physical and socioeconomic environments of cities},
  author={Hou, Ce and Zhang, Fan and Kang, Yuhao and Gao, Song and Li, Yong and Duarte, F{\'a}bio and Li, Sen},
  journal={Annals of the American Association of Geographers},
  volume={115},
  number={1},
  pages={148--166},
  year={2025},
  publisher={Taylor \& Francis}
}

@article{yu2025exploring,
  title={Exploring geo-transferability of deep neural network by developing comprehensive metrics},
  author={Yu, Junxian and Tu, Wei and Chen, Xiana and Li, Mingxiao and Chen, Dongsheng and Cao, Rui and Li, Qingquan},
  journal={Geo-spatial Information Science},
  pages={1--19},
  year={2025},
  publisher={Taylor \& Francis}
}

@misc{liu2023towards,
      title={Towards Out-Of-Distribution Generalization: A Survey}, 
      author={Jiashuo Liu and Zheyan Shen and Yue He and Xingxuan Zhang and Renzhe Xu and Han Yu and Peng Cui},
      year={2023},
      eprint={2108.13624},
      archivePrefix={arXiv},
      primaryClass={cs.LG},
      url={https://arxiv.org/abs/2108.13624}, 
}

@article{zhang2026city,
  title={City identity recognition: how representation bias influences model predictability and replicability?},
  author={Zhang, Xiang and Yang, Fan and He, Zongze and Li, Weijia and Yang, Min},
  journal={Computers, Environment and Urban Systems},
  volume={123},
  pages={102370},
  year={2026},
  publisher={Elsevier}
}

@article{simini2021deep,
  title={A deep gravity model for mobility flows generation},
  author={Simini, Filippo and Barlacchi, Gianni and Luca, Massimilano and Pappalardo, Luca},
  journal={Nature communications},
  volume={12},
  number={1},
  pages={6576},
  year={2021},
  publisher={Nature Publishing Group UK London}
}

@inproceedings{rong2025large,
 author = {Rong, Can and Ding, Jingtao and Liu, Yan and Li, Yong},
 booktitle = {International Conference on Representation Learning},
 editor = {Y. Yue and A. Garg and N. Peng and F. Sha and R. Yu},
 pages = {96180--96205},
 title = {A Large-scale Dataset and Benchmark for Commuting Origin-Destination Flow Generation},
 volume = {2025},
 year = {2025}
}

@article{wang2025geobs,
  title={GeoBS: Information-Theoretic Quantification of Geographic Bias in AI Models},
  author={Wang, Zhangyu and Wu, Nemin and Cao, Qian and Xia, Jiangnan and Liu, Zeping and Xie, Yiqun and Nambi, Akshay and Ganu, Tanuja and Lao, Ni and Liu, Ninghao and others},
  journal={arXiv preprint arXiv:2509.23482},
  year={2025}
}

@article{chen2025addressing,
  title={Addressing the fixed effects in gravity model based on higher-order origin-destination pairs},
  author={Chen, Yuzhou and Ma, Qiwei and Tao, Ran},
  journal={International Journal of Geographical Information Science},
  volume={39},
  number={5},
  pages={1162--1182},
  year={2025},
  publisher={Taylor \& Francis}
}

@article{guo2025regiongcn,
  title={RegionGCN: Spatial-heterogeneity-aware graph convolutional networks},
  author={Guo, Hao and Wang, Han and Zhu, Di and Wu, Lun and Fotheringham, A Stewart and Liu, Yu},
  journal={Annals of the American Association of Geographers},
  pages={1--17},
  year={2025},
  publisher={Taylor \& Francis}
}

@article{xu2025predicting,
  title={Predicting human mobility flows in cities using deep learning on satellite imagery},
  author={Xu, Yichen and Gao, Song and Huang, Qunying and G{\"o}{\c{c}}men, Asl{\i}g{\"u}l and Zhu, Qiang and Zhang, Feng},
  journal={Nature Communications},
  volume={16},
  number={1},
  pages={10372},
  year={2025},
  publisher={Nature Publishing Group UK London}
}

@article{boucherie2025decoupling,
  title={Decoupling geographical constraints from human mobility},
  author={Boucherie, Louis and Maier, Benjamin F and Lehmann, Sune},
  journal={Nature Human Behaviour},
  pages={1--12},
  year={2025},
  publisher={Nature Publishing Group UK London}
}

@article{luca2021survey,
  title={A survey on deep learning for human mobility},
  author={Luca, Massimiliano and Barlacchi, Gianni and Lepri, Bruno and Pappalardo, Luca},
  journal={ACM Computing Surveys (CSUR)},
  volume={55},
  number={1},
  pages={1--44},
  year={2021},
  publisher={ACM New York, NY}
}

@article{xu2025using,
  title={Using human mobility data to quantify experienced urban inequalities},
  author={Xu, Fengli and Wang, Qi and Moro, Esteban and Chen, Lin and Salazar Miranda, Arianna and Gonz{\'a}lez, Marta C and Tizzoni, Michele and Song, Chaoming and Ratti, Carlo and Bettencourt, Luis and others},
  journal={Nature Human Behaviour},
  pages={1--11},
  year={2025},
  publisher={Nature Publishing Group UK London}
}

@article{santana2023covid,
  title={COVID-19 is linked to changes in the time--space dimension of human mobility},
  author={Santana, Clodomir and Botta, Federico and Barbosa, Hugo and Privitera, Filippo and Menezes, Ronaldo and Di Clemente, Riccardo},
  journal={Nature Human Behaviour},
  volume={7},
  number={10},
  pages={1729--1739},
  year={2023},
  publisher={Nature Publishing Group UK London}
}

@article{yabe2025behaviour,
  title={Behaviour-based dependency networks between places shape urban economic resilience},
  author={Yabe, Takahiro and Garc{\'\i}a Bulle Bueno, Bernardo and Frank, Morgan R and Pentland, Alex and Moro, Esteban},
  journal={Nature human behaviour},
  volume={9},
  number={3},
  pages={496--506},
  year={2025},
  publisher={Nature Publishing Group UK London}
}

@article{wang2024infrequent,
  title={Infrequent activities predict economic outcomes in major American cities},
  author={Wang, Shenhao and Zheng, Yunhan and Wang, Guang and Yabe, Takahiro and Moro, Esteban and Pentland, Alex ‘Sandy’},
  journal={Nature Cities},
  volume={1},
  number={4},
  pages={305--314},
  year={2024},
  publisher={Nature Publishing Group US New York}
}

@article{nilforoshan2023human,
  title={Human mobility networks reveal increased segregation in large cities},
  author={Nilforoshan, Hamed and Looi, Wenli and Pierson, Emma and Villanueva, Blanca and Fishman, Nic and Chen, Yiling and Sholar, John and Redbird, Beth and Grusky, David and Leskovec, Jure},
  journal={Nature},
  volume={624},
  number={7992},
  pages={586--592},
  year={2023},
  publisher={Nature Publishing Group UK London}
}

@article{zheng2024impacts,
  title={Impacts of remote work on vehicle miles traveled and transit ridership in the USA},
  author={Zheng, Yunhan and Wang, Shenhao and Liu, Lun and Aloisi, Jim and Zhao, Jinhua},
  journal={Nature Cities},
  volume={1},
  number={5},
  pages={346--358},
  year={2024},
  publisher={Nature Publishing Group US New York}
}

@article{pappalardo2023future,
  title={Future directions in human mobility science},
  author={Pappalardo, Luca and Manley, Ed and Sekara, Vedran and Alessandretti, Laura},
  journal={Nature computational science},
  volume={3},
  number={7},
  pages={588--600},
  year={2023},
  publisher={Nature Publishing Group US New York}
}

@article{gonzalez2008understanding,
  title={Understanding individual human mobility patterns},
  author={Gonzalez, Marta C and Hidalgo, Cesar A and Barabasi, Albert-Laszlo},
  journal={nature},
  volume={453},
  number={7196},
  pages={779--782},
  year={2008},
  publisher={Nature Publishing Group UK London}
}

@article{schlapfer2021universal,
  title={The universal visitation law of human mobility},
  author={Schl{\"a}pfer, Markus and Dong, Lei and O’Keeffe, Kevin and Santi, Paolo and Szell, Michael and Salat, Hadrien and Anklesaria, Samuel and Vazifeh, Mohammad and Ratti, Carlo and West, Geoffrey B},
  journal={Nature},
  volume={593},
  number={7860},
  pages={522--527},
  year={2021},
  publisher={Nature Publishing Group UK London}
}

@article{xu2023urban,
  title={Urban dynamics through the lens of human mobility},
  author={Xu, Yanyan and Olmos, Luis E and Mateo, David and Hernando, Alberto and Yang, Xiaokang and Gonz{\'a}lez, Marta C},
  journal={Nature Computational Science},
  volume={3},
  number={7},
  pages={611--620},
  year={2023},
  publisher={Nature Publishing Group US New York}
}

@article{rong2023goddag,
  title={{GODDAG}: Generating origin-destination flow for new cities via domain adversarial training},
  author={Rong, Can and Feng, Jie and Ding, Jingtao},
  journal={IEEE Transactions on Knowledge and Data Engineering},
  volume={35},
  number={10},
  pages={10048--10057},
  year={2023},
  publisher={IEEE}
}

@article{rong2023origin,
  title={Origin-destination network generation via gravity-guided {GAN}},
  author={Rong, Can and Wang, Huandong and Li, Yong},
  journal={arXiv preprint arXiv:2306.03390},
  year={2023}
}

@inproceedings{wang2025deep,
  title={A Deep Origin-Destination Flow Imputation Model Informed by the Visitation Law in Human Mobility},
  author={Wang, Sheng and Zhu, Di},
  booktitle={Proceedings of the 33rd ACM International Conference on Advances in Geographic Information Systems},
  pages={1266--1269},
  year={2025}
}

@article{zhao2025predicting,
  title={Predicting origin-destination flows by considering heterogeneous mobility patterns},
  author={Zhao, Yibo and Cheng, Shifen and Gao, Song and Wang, Peixiao and Lu, Feng},
  journal={Sustainable Cities and Society},
  volume={118},
  pages={106015},
  year={2025},
  publisher={Elsevier}
}

@article{tobler1970computer,
  title={A computer movie simulating urban growth in the Detroit region},
  author={Tobler, Waldo R},
  journal={Economic geography},
  volume={46},
  number={sup1},
  pages={234--240},
  year={1970},
  publisher={Taylor \& Francis}
}

@article{goodchild2004validity,
  title={The validity and usefulness of laws in geographic information science and geography},
  author={Goodchild, Michael F},
  journal={Annals of the Association of American Geographers},
  volume={94},
  number={2},
  pages={300--303},
  year={2004},
  publisher={Taylor \& Francis}
}

@article{zhu2018spatial,
  title={Spatial prediction based on Third Law of Geography},
  author={Zhu, A-Xing and Lu, Guonian and Liu, Jing and Qin, Cheng-Zhi and Zhou, Chenghu},
  journal={Annals of GIS},
  volume={24},
  number={4},
  pages={225--240},
  year={2018},
  publisher={Taylor \& Francis}
}

@article{zhang2024urban,
  title={Urban visual intelligence: Studying cities with artificial intelligence and street-level imagery},
  author={Zhang, Fan and Salazar-Miranda, Arianna and Duarte, F{\'a}bio and Vale, Lawrence and Hack, Gary and Chen, Min and Liu, Yu and Batty, Michael and Ratti, Carlo},
  journal={Annals of the American Association of Geographers},
  volume={114},
  number={5},
  pages={876--897},
  year={2024},
  publisher={Taylor \& Francis}
}

@article{zhao2025multivariate,
  title={A multivariate spatial structure indicator based on geographic similarity},
  author={Zhao, Fang-He and Qin, Cheng-Zhi and Zhu, A-Xing and Pei, Tao},
  journal={International Journal of Geographical Information Science},
  volume={39},
  number={7},
  pages={1518--1539},
  year={2025},
  publisher={Taylor \& Francis}
}

@article{yan2024quantifying,
  title={Quantifying spatial similarity for use as constraints in map generalisation},
  author={Yan, Haowen},
  journal={Journal of Spatial Science},
  volume={69},
  number={1},
  pages={23--42},
  year={2024},
  publisher={Taylor \& Francis}
}

@article{xu2023spatial,
  title={Spatial patterns and determinants of inter-county migration in California: A multilevel gravity model approach},
  author={Xu, Anqi},
  journal={Population Research and Policy Review},
  volume={42},
  number={3},
  pages={40},
  year={2023},
  publisher={Springer}
}

@article{long2025data,
  title={Data-driven movement analysis},
  author={Long, JA and Dem{\v{s}}ar, Ur{\v{s}}ka and Dodge, S and Weibel, Robert},
  journal={International Journal of Geographical Information Science},
  volume={39},
  number={5},
  pages={945--950},
  year={2025},
  publisher={Taylor \& Francis}
}

@article{lou2025geoxcp,
  title={GeoXCP: uncertainty quantification of spatial explanations in explainable AI},
  author={Lou, Xiayin and Luo, Peng and Li, Ziqi and Gao, Song and Meng, Liqiu},
  journal={International Journal of Geographical Information Science},
  pages={1--31},
  year={2025},
  publisher={Taylor \& Francis}
}

@article{zhuang2020comprehensive,
  title={A comprehensive survey on transfer learning},
  author={Zhuang, Fuzhen and Qi, Zhiyuan and Duan, Keyu and Xi, Dongbo and Zhu, Yongchun and Zhu, Hengshu and Xiong, Hui and He, Qing},
  journal={Proceedings of the IEEE},
  volume={109},
  number={1},
  pages={43--76},
  year={2020},
  publisher={Ieee}
}

@article{barbosa2018human,
  title={Human mobility: Models and applications},
  author={Barbosa, Hugo and Barthelemy, Marc and Ghoshal, Gourab and James, Charlotte R and Lenormand, Maxime and Louail, Thomas and Menezes, Ronaldo and Ramasco, Jos{\'e} J and Simini, Filippo and Tomasini, Marcello},
  journal={Physics Reports},
  volume={734},
  pages={1--74},
  year={2018},
  publisher={Elsevier}
}

@inproceedings{he2020human,
  title={What is the human mobility in a new city: Transfer mobility knowledge across cities},
  author={He, Tianfu and Bao, Jie and Li, Ruiyuan and Ruan, Sijie and Li, Yanhua and Song, Li and He, Hui and Zheng, Yu},
  booktitle={Proceedings of The Web Conference 2020},
  pages={1355--1365},
  year={2020}
}

@inproceedings{wang2024cola,
  title={Cola: Cross-city mobility transformer for human trajectory simulation},
  author={Wang, Yu and Zheng, Tongya and Liang, Yuxuan and Liu, Shunyu and Song, Mingli},
  booktitle={Proceedings of the ACM Web Conference 2024},
  pages={3509--3520},
  year={2024}
}

@article{jiang2021transfer,
  title={Transfer urban human mobility via poi embedding over multiple cities},
  author={Jiang, Renhe and Song, Xuan and Fan, Zipei and Xia, Tianqi and Wang, Zhaonan and Chen, Quanjun and Cai, Zekun and Shibasaki, Ryosuke},
  journal={ACM Transactions on Data Science},
  volume={2},
  number={1},
  pages={1--26},
  year={2021},
  publisher={ACM New York, NY, USA}
}

@inproceedings{yu2024harnessing,
  title={Harnessing llms for cross-city od flow prediction},
  author={Yu, Chenyang and Xie, Xinpeng and Huang, Yan and Qiu, Chenxi},
  booktitle={Proceedings of the 32nd ACM International Conference on Advances in Geographic Information Systems},
  pages={384--395},
  year={2024}
}

@article{mauro2022generating,
  title={Generating mobility networks with generative adversarial networks},
  author={Mauro, Giovanni and Luca, Massimiliano and Longa, Antonio and Lepri, Bruno and Pappalardo, Luca},
  journal={EPJ data science},
  volume={11},
  number={1},
  pages={58},
  year={2022},
  publisher={Springer Berlin Heidelberg}
}

@article{yuan2025learning,
  title={Learning the complexity of urban mobility with deep generative network},
  author={Yuan, Yuan and Ding, Jingtao and Jin, Depeng and Li, Yong},
  journal={PNAS nexus},
  volume={4},
  number={5},
  pages={pgaf081},
  year={2025},
  publisher={Oxford University Press US}
}

@article{pan2009survey,
  title={A survey on transfer learning},
  author={Pan, Sinno Jialin and Yang, Qiang},
  journal={IEEE Transactions on knowledge and data engineering},
  volume={22},
  number={10},
  pages={1345--1359},
  year={2009},
  publisher={IEEE}
}

@article{cai2025diagnosing,
  title={Diagnosing model performance under distribution shift},
  author={Cai, Tiffany and Namkoong, Hongseok and Yadlowsky, Steve},
  journal={Operations Research},
  year={2025},
  publisher={INFORMS}
}

@article{garg2020unified,
  title={A unified view of label shift estimation},
  author={Garg, Saurabh and Wu, Yifan and Balakrishnan, Sivaraman and Lipton, Zachary},
  journal={Advances in Neural Information Processing Systems},
  volume={33},
  pages={3290--3300},
  year={2020}
}

@article{moreno2012unifying,
  title={A unifying view on dataset shift in classification},
  author={Moreno-Torres, Jose G and Raeder, Troy and Alaiz-Rodr{\'\i}guez, Roc{\'\i}o and Chawla, Nitesh V and Herrera, Francisco},
  journal={Pattern recognition},
  volume={45},
  number={1},
  pages={521--530},
  year={2012},
  publisher={Elsevier}
}

@inproceedings{blitzer2007biographies,
  title={Biographies, bollywood, boom-boxes and blenders: Domain adaptation for sentiment classification},
  author={Blitzer, John and Dredze, Mark and Pereira, Fernando},
  booktitle={Proceedings of the 45th annual meeting of the association of computational linguistics},
  pages={440--447},
  year={2007}
}

@article{farahani2021brief,
  title={A brief review of domain adaptation},
  author={Farahani, Abolfazl and Voghoei, Sahar and Rasheed, Khaled and Arabnia, Hamid R},
  journal={Advances in data science and information engineering: proceedings from ICDATA 2020 and IKE 2020},
  pages={877--894},
  year={2021},
  publisher={Springer}
}

@article{gretton2006kernel,
  title={A kernel method for the two-sample-problem},
  author={Gretton, Arthur and Borgwardt, Karsten and Rasch, Malte and Sch{\"o}lkopf, Bernhard and Smola, Alex},
  journal={Advances in neural information processing systems},
  volume={19},
  year={2006}
}

@article{kullback1951information,
  title={On information and sufficiency},
  author={Kullback, Solomon and Leibler, Richard A},
  journal={The annals of mathematical statistics},
  volume={22},
  number={1},
  pages={79--86},
  year={1951},
  publisher={JSTOR}
}

@incollection{sun2017correlation,
  title={Correlation alignment for unsupervised domain adaptation},
  author={Sun, Baochen and Feng, Jiashi and Saenko, Kate},
  booktitle={Domain adaptation in computer vision applications},
  pages={153--171},
  year={2017},
  publisher={Springer}
}

@article{panaretos2019statistical,
  title={Statistical aspects of Wasserstein distances},
  author={Panaretos, Victor M and Zemel, Yoav},
  journal={Annual review of statistics and its application},
  volume={6},
  number={1},
  pages={405--431},
  year={2019},
  publisher={Annual Reviews}
}

@article{janowicz2011semantics,
  title={The semantics of similarity in geographic information retrieval},
  author={Janowicz, Krzysztof and Raubal, Martin and Kuhn, Werner},
  journal={Journal of Spatial Information Science},
  volume={2},
  pages={29--57},
  year={2011},
  publisher={Univ. of Maine}
}

@article{mcintosh2005assessing,
  title={Assessing similarity of geographic processes and events},
  author={McIntosh, John and Yuan, May},
  journal={Transactions in GIS},
  volume={9},
  number={2},
  pages={223--245},
  year={2005},
  publisher={Wiley Online Library}
}

@article{moran1950notes,
  title={Notes on continuous stochastic phenomena},
  author={Moran, Patrick AP},
  journal={Biometrika},
  volume={37},
  number={1/2},
  pages={17--23},
  year={1950},
  publisher={JSTOR}
}

@article{zhu2015predictive,
  title={Predictive soil mapping with limited sample data},
  author={Zhu, AX and Liu, J and Du, F and Zhang, SJ and Burt, J and Behrens, T and Scholten, T},
  journal={European Journal of Soil Science},
  volume={66},
  number={3},
  pages={535--547},
  year={2015},
  publisher={Wiley Online Library}
}

@article{schwering2008approaches,
  title={Approaches to semantic similarity measurement for geo-spatial data: a survey},
  author={Schwering, Angela},
  journal={Transactions in GIS},
  volume={12},
  number={1},
  pages={5--29},
  year={2008},
  publisher={Wiley Online Library}
}

@article{lee2017extending,
  title={Extending moran's index for measuring spatiotemporal clustering of geographic events},
  author={Lee, Jay and Li, Shengwen},
  journal={Geographical Analysis},
  volume={49},
  number={1},
  pages={36--57},
  year={2017},
  publisher={Wiley Online Library}
}

@article{goodchild2021replication,
  title={Replication across space and time must be weak in the social and environmental sciences},
  author={Goodchild, Michael F and Li, Wenwen},
  journal={Proceedings of the National Academy of Sciences},
  volume={118},
  number={35},
  pages={e2015759118},
  year={2021},
  publisher={National Academy of Sciences}
}

@article{lin2023comparison,
  title={{Comparison of Moran's I and Geary's C in multivariate spatial pattern analysis}},
  author={Lin, Jie},
  journal={Geographical Analysis},
  volume={55},
  number={4},
  pages={685--702},
  year={2023},
  publisher={Wiley Online Library}
}

@article{liu2025generating,
  title={Generating equitable urban human flows with a fairness-aware deep learning model},
  author={Liu, Zhewei and Huang, Lipai and Fan, Chao and Mostafavi, Ali},
  journal={Cities},
  volume={167},
  pages={106296},
  year={2025},
  publisher={Elsevier}
}

@article{schlosser2021biases,
  title={Biases in human mobility data impact epidemic modeling},
  author={Schlosser, Frank and Sekara, Vedran and Brockmann, Dirk and Garcia-Herranz, Manuel},
  journal={arXiv preprint arXiv:2112.12521},
  year={2021}
}

@article{nijs2025data,
  title={Data Bias in Human Mobility is a Universal Phenomenon but is Highly Location-specific},
  author={Nijs, Katinka den and Omodei, Elisa and Sekara, Vedran},
  journal={arXiv preprint arXiv:2508.00149},
  year={2025}
}

@article{merchant2023scaling,
  title={Scaling deep learning for materials discovery},
  author={Merchant, Amil and Batzner, Simon and Schoenholz, Samuel S and Aykol, Muratahan and Cheon, Gowoon and Cubuk, Ekin Dogus},
  journal={Nature},
  volume={624},
  number={7990},
  pages={80--85},
  year={2023},
  publisher={Nature Publishing Group UK London}
}

@inproceedings{cherti2023reproducible,
  title={Reproducible scaling laws for contrastive language-image learning},
  author={Cherti, Mehdi and Beaumont, Romain and Wightman, Ross and Wortsman, Mitchell and Ilharco, Gabriel and Gordon, Cade and Schuhmann, Christoph and Schmidt, Ludwig and Jitsev, Jenia},
  booktitle={Proceedings of the IEEE/CVF conference on computer vision and pattern recognition},
  pages={2818--2829},
  year={2023}
}

@article{hestness2017deep,
  title={Deep learning scaling is predictable, empirically},
  author={Hestness, Joel and Narang, Sharan and Ardalani, Newsha and Diamos, Gregory and Jun, Heewoo and Kianinejad, Hassan and Patwary, Md Mostofa Ali and Yang, Yang and Zhou, Yanqi},
  journal={arXiv preprint arXiv:1712.00409},
  year={2017}
}

@inproceedings{li2025cross,
  title={Cross City Traffic Flow Generation via Retrieval Augmented Diffusion Model},
  author={Li, Yudong and Wang, Jingyuan and Yu, Xie and Wang, Peiyu and Huang, Qian},
  booktitle={The Thirty-ninth Annual Conference on Neural Information Processing Systems},
  year={2025}
}

@article{rong2024interdisciplinary,
  title={An interdisciplinary survey on origin-destination flows modeling: Theory and techniques},
  author={Rong, Can and Ding, Jingtao and Li, Yong},
  journal={ACM Computing Surveys},
  volume={57},
  number={1},
  pages={1--49},
  year={2024},
  publisher={ACM New York, NY}
}

@inproceedings{deng2025geoaggregator,
  title={GeoAggregator: An efficient transformer model for geo-spatial tabular data},
  author={Deng, Rui and Li, Ziqi and Wang, Mingshu},
  booktitle={Proceedings of the AAAI Conference on Artificial Intelligence},
  volume={39},
  number={11},
  pages={11572--11580},
  year={2025}
}

@article{janowicz2025geofm,
  title={GeoFM: how will geo-foundation models reshape spatial data science and GeoAI?},
  author={Janowicz, Krzysztof and Mai, Gengchen and Huang, Weiming and Zhu, Rui and Lao, Ni and Cai, Ling},
  journal={International Journal of Geographical Information Science},
  volume={39},
  number={9},
  pages={1849--1865},
  year={2025},
  publisher={Taylor \& Francis}
}

@inproceedings{kalluri2023geonet,
  title={Geonet: Benchmarking unsupervised adaptation across geographies},
  author={Kalluri, Tarun and Xu, Wangdong and Chandraker, Manmohan},
  booktitle={Proceedings of the IEEE/CVF Conference on Computer Vision and Pattern Recognition},
  pages={15368--15379},
  year={2023}
}

@article{al2025analysing,
  title={Analysing Satellite Imagery Classification under Spatial Domain Shift across Geographic Regions},
  author={Al-Emadi, Sara A and Yang, Yin and Ofli, Ferda},
  journal={International Journal of Computer Vision},
  volume={133},
  number={11},
  pages={7672--7709},
  year={2025},
  publisher={Springer}
}

@inproceedings{al2025benchmarking,
  title={Benchmarking Object Detectors under Real-World Distribution Shifts in Satellite Imagery},
  author={Al-Emadi, Sara A and Yang, Yin and Ofli, Ferda},
  booktitle={Proceedings of the Computer Vision and Pattern Recognition Conference},
  pages={8299--8309},
  year={2025}
}

@article{crasto2025robustness,
  title={Robustness to Geographic Distribution Shift using Location Encoders},
  author={Crasto, Ruth},
  journal={arXiv preprint arXiv:2503.02036},
  year={2025}
}

@article{stewart2025torchgeo,
  title={Torchgeo: deep learning with geospatial data},
  author={Stewart, Adam J and Robinson, Caleb and Corley, Isaac A and Ortiz, Anthony and Lavista Ferres, Juan M and Banerjee, Arindam},
  journal={ACM Transactions on Spatial Algorithms and Systems},
  volume={11},
  number={4},
  pages={1--28},
  year={2025},
  publisher={ACM New York, NY}
}

@inproceedings{zhao2024measuring,
  title={Measuring diversity in datasets},
  author={Zhao, Dorothy and Andrews, Jerone TA and Sony, AI and Papakyriakopoulos, Tokyo Orestis and Xiang, Alice},
  booktitle={International Conference on Learning Representations},
  volume={1},
  pages={36},
  year={2024}
}

@article{aroyo2023dices,
  title={Dices dataset: Diversity in conversational ai evaluation for safety},
  author={Aroyo, Lora and Taylor, Alex and Diaz, Mark and Homan, Christopher and Parrish, Alicia and Serapio-Garc{\'\i}a, Gregory and Prabhakaran, Vinodkumar and Wang, Ding},
  journal={Advances in Neural Information Processing Systems},
  volume={36},
  pages={53330--53342},
  year={2023}
}

@inproceedings{mandal2021dataset,
  title={Dataset diversity: measuring and mitigating geographical bias in image search and retrieval},
  author={Mandal, Abhishek and Leavy, Susan and Little, Suzanne},
  booktitle={Proceedings of the 1st International Workshop on Trustworthy AI for Multimedia Computing},
  pages={19--25},
  year={2021}
}

@inproceedings{nougnanke2025dataset,
  title={How Dataset Diversity Affects Generalization in ML-based NIDS},
  author={Nougnanke, Benoit and Blanc, Gregory and Robert, Thomas},
  booktitle={European Symposium on Research in Computer Security},
  pages={269--288},
  year={2025},
  organization={Springer}
}

@article{zhu2022third,
  title={How is the third law of geography different?},
  author={Zhu, A-Xing and Turner, Matthew},
  journal={Annals of GIS},
  volume={28},
  number={1},
  pages={57--67},
  year={2022},
  publisher={Taylor \& Francis}
}

@inproceedings{moska2025obsr,
  title={OBSR: Open Benchmark for Spatial Representations},
  author={Moska, Julia and Furman, Oleksii and Kozaczko, Kacper and Leszkiewicz, Szymon and Polczyk, Jakub and Gramacki, Piotr and Szyma{\'n}ski, Piotr},
  booktitle={Proceedings of the 33rd ACM International Conference on Advances in Geographic Information Systems},
  pages={670--681},
  year={2025}
}

@article{pourebrahim2019trip,
  title={Trip distribution modeling with Twitter data},
  author={Pourebrahim, Nastaran and Sultana, Selima and Niakanlahiji, Amirreza and Thill, Jean-Claude},
  journal={Computers, Environment and Urban Systems},
  volume={77},
  pages={101354},
  year={2019},
  publisher={Elsevier}
}

@inproceedings{robinson2018machine,
  title={A machine learning approach to modeling human migration},
  author={Robinson, Caleb and Dilkina, Bistra},
  booktitle={Proceedings of the 1st ACM SIGCAS Conference on Computing and Sustainable Societies},
  pages={1--8},
  year={2018}
}

@inproceedings{liu2020learning,
  title={Learning geo-contextual embeddings for commuting flow prediction},
  author={Liu, Zhicheng and Miranda, Fabio and Xiong, Weiting and Yang, Junyan and Wang, Qiao and Silva, Claudio},
  booktitle={Proceedings of the AAAI conference on artificial intelligence},
  volume={34},
  number={01},
  pages={808--816},
  year={2020}
}

\end{document}